\documentclass[10pt,twocolumn,letterpaper]{article}

\usepackage{iccv}              % To produce the CAMERA-READY version
\usepackage[accsupp]{axessibility}
\usepackage{booktabs}
\usepackage{xcolor}
\usepackage{graphicx}
\usepackage{caption}
\usepackage{subcaption}
\usepackage{amsmath}
\usepackage[english]{babel}
\usepackage{amsfonts}
\usepackage{mathtools, nccmath}
\usepackage[linesnumbered,ruled,vlined]{algorithm2e}
\usepackage{multirow}
\usepackage{capt-of}
\definecolor{iccvblue}{rgb}{0.21,0.49,0.74}
\usepackage[pagebackref,breaklinks,colorlinks,allcolors=iccvblue]{hyperref}

\def\paperID{14209} % *** Enter the Paper ID here
\def\confName{ICCV}
\def\confYear{2025}

\title{ConstStyle:  Robust Domain Generalization with Unified Style Transformation}

\author{
Nam Duong Tran \textsuperscript{1} \quad 
Nam Nguyen Phuong \textsuperscript{1} \quad 
Hieu H. Pham \textsuperscript{2} \quad 
Phi Le Nguyen \textsuperscript{1}\thanks{Corresponding Authors} \quad 
My T. Thai \textsuperscript{3}\footnotemark[1] \\
\textsuperscript{1} Institute for AI Innovation and Societal Impact, Hanoi University of Science and Technology \\ \textsuperscript{2} VinUniversity \quad \textsuperscript{3} University of Florida\\
}
\newcommand{\nduong}[1]
{\textcolor{blue}{#1}}

\newcommand{\nam}[1]{\textbf{\textcolor{purple}{#1}}}

\newtheorem{theorem}{Theorem}

\newtheorem{lemma}{Lemma}
\newtheorem{definition}{Definition}

\begin{document}
\maketitle
\begin{abstract}
Deep neural networks often suffer performance drops when test data distribution differs from training data. Domain Generalization (DG) aims to address this by focusing on domain-invariant features or augmenting data for greater diversity. However, these methods often struggle with limited training domains or significant gaps between seen (training) and unseen (test) domains. To enhance DG robustness, we hypothesize that it is essential for the model to be trained on data from domains that closely resemble unseen test domains—an inherently difficult task due to the absence of prior knowledge about the unseen domains. Accordingly, we propose ConstStyle, a novel approach that leverages a unified domain to capture domain-invariant features and bridge the domain gap with theoretical analysis. During training, all samples are mapped onto this unified domain, optimized for seen domains. During testing, unseen domain samples are projected similarly before predictions. By aligning both training and testing data within this unified domain, ConstStyle effectively reduces the impact of domain shifts, even with large domain gaps or few seen domains. Extensive experiments demonstrate that ConstStyle consistently outperforms existing methods across diverse scenarios. Notably, when only a limited number of seen domains are available, ConstStyle can boost accuracy up to 19.82\% compared to the next best approach. \footnote{Source code: \url{https://github.com/nduongw/ConstStyle}}
\end{abstract}    
\vspace{-12pt}
\section{Introduction}
\label{sec:intro}
%When deploying models trained on specific datasets (i.e., seen domain) to data with differing distributions (i.e., unseen domain), model performance often suffers substantially degradation due to domain shift. Given the extensive diversity of real-world data, collecting data that represents all domains for comprehensive model training is impractical.  As a result, closing the domain gap between training and testing data remains a major challenge.

Deep neural networks (DNNs) often experience significant performance degradation when deployed on unseen test domains, which differ in distribution from the training data. This issue, known as domain shift, poses a fundamental challenge in real-world applications where data distributions are inherently diverse and unpredictable. Since it is impractical to collect data representative of all possible domains, bridging the gap between training (seen) and testing (unseen) domains is crucial for achieving robust performance—yet remains a major hurdle.

\begin{figure*}
    \centering
    \begin{minipage}{0.35\linewidth} 
        \begin{subfigure}[b]{0.45\columnwidth}
        \includegraphics[width=\columnwidth]{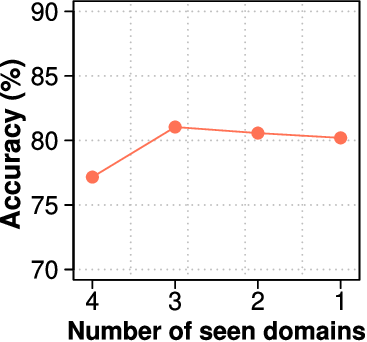}
        \caption{Digits5 dataset.}
        \label{fig:intro-digits}
        \end{subfigure}
        \hfill
        \begin{subfigure}[b]{0.45\columnwidth}
            \includegraphics[width=\columnwidth]{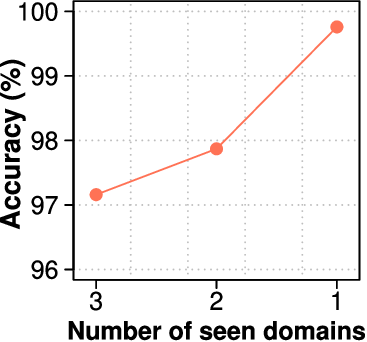}
            \caption{VLCS dataset.}
            \label{fig:intro-vlcs}
        \end{subfigure}
        \caption{Accuracy of ConvNet with varying numbers of seen domains during training. In some cases, using fewer seen domains leads to improved performance.}
        \label{fig:accuracy drop}
        % \caption{Accuracy of ConvNet with varying numbers of seen domains during training. For the Digits5 dataset, the test domain is SVHN, and the seen domains are (SYN, USPS, MNIST, MNISTM), (SYN, USPS, MNIST), (SYN, USPS), and (SYN). For the VLCS dataset, the test domain is Caltech, with the seen being (Pascal, Sun, Labelme), (Pascal, Sun), and (Pascal).}
    \end{minipage}
    \hfill
    \begin{minipage}{0.58\linewidth} 
        \begin{subfigure}[b]{0.45\columnwidth}
        \includegraphics[width=1\columnwidth]{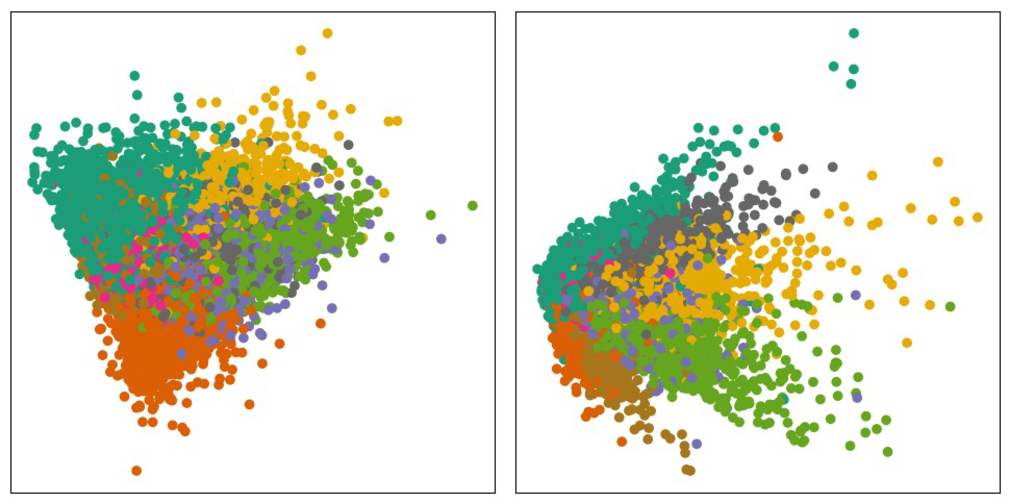}
        \caption{Training ConvNet with 4 seen domains (left) vs single seen domain (right) on Digits5 dataset.}
        \label{fig:feats-digits}
        \end{subfigure}
        \hspace{5pt}
        \begin{subfigure}[b]{0.45\columnwidth}
            \includegraphics[width=1\columnwidth]{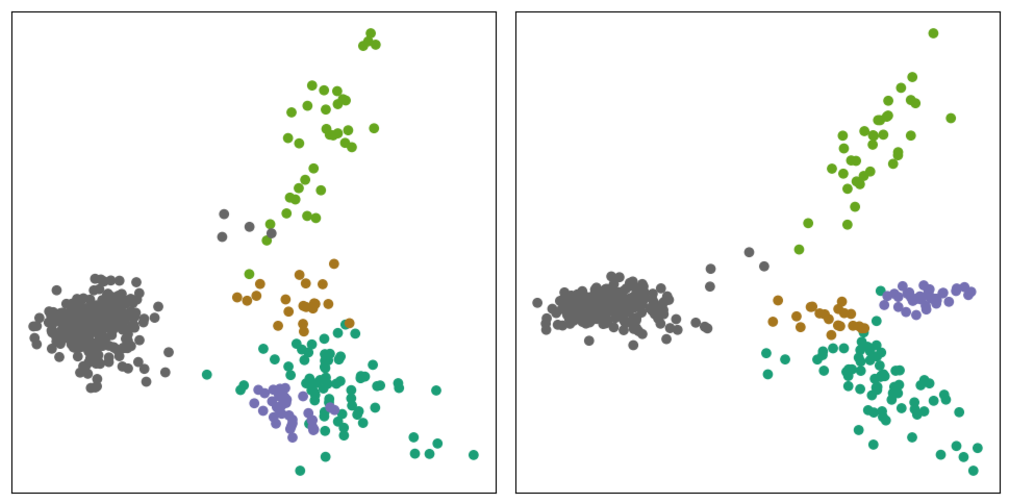}
            \caption{Training ResNet18 with  3 seen domains (left) vs single seen domain (right) on VLCS dataset.}
            \label{fig:feats-vlcs}
        \end{subfigure}
        \caption{The illustrations indicate that training on a single seen domain can sometimes produce more defined class boundaries than training across multiple domains. Each dot represents an instance, with colors signifying the labels.}
        \label{fig:visual drop}
    \end{minipage}
    \vspace{-10pt}
\end{figure*}

Domain generalization (DG) addresses this issue by training models that generalize well to unseen domains without relying on their data during training \cite{zhou2022domain, wang2022generalizing}. %improving the robustness of deep neural networks (DNN) across various domains \cite{zhou2022domain, wang2022generalizing}. 
Existing DG methods primarily focus on two strategies: (1) learning domain-invariant features, and (2) augmenting training data to enhance domain diversity. Invariant representation learning methods extract shared features across domains, minimizing the impact of domain-specific variations \cite{choi2021robustnet, li2018deep, chen2023domain, piratla2020efficient}. However, these approaches typically require numerous diverse domains to effectively capture invariance, making them costly and often impractical. Alternatively, data augmentation methods increase domain diversity \cite{zhou2021domain, nuriel2021permuted, li2022uncertainty, zhang2024domain}, fundamentally based on the assumption that training with more data from diverse domains will produce better performance.
Yet, our empirical analysis shows that increasing the number of seen domains does not always improve performance on unseen domains. In fact, training on fewer but carefully selected domains can sometimes yield better generalization, such as resulting in higher accuracy, as shown in Figure \ref{fig:accuracy drop} or having greater class separation, as shown in Figure \ref{fig:visual drop}.

Furthermore, existing methods typically emphasize the training process, focusing on seen domains while neglecting the testing phase, where the domain gap becomes most pronounced. This oversight reveals another key limitation in current approaches, as they do not adequately address the unseen domain at inference time. This leads to a critical question: How can we enhance model generalization on unseen domains that are unknown during training but must be handled effectively during inference?

To overcome these challenges, we propose ConstStyle, a novel DG framework that unifies the treatment of both training and testing processes. %That is, we consider both training and testing processes to simultaneously achieve two goals: learning domain-invariant features from seen domains during training and reducing the domain gap between unseen and seen domains during testing, thus boosting the model’s performance on generalization.  
ConstStyle leverages our newly introduced concept of \textit{unified domain}, which acts as a common representation space that minimizes style discrepancies between different domains, both seen and unseen. Specifically, during training, all samples are aligned with this unified domain to extract consistent features. At inference, test samples undergo a style transformation to match the unified domain. By aligning data in this manner, ConstStyle reduces the impact of domain shifts, even when there is a significant gap between seen and unseen domains. Our approach is grounded in a theoretical framework that guides the selection of the unified domain to optimize generalization, providing a robust solution to the DG problem. %onto which all data—both seen and unseen—is projected prior to both training and testing. This unified domain

The main contributions of this paper are as follows: 
\begin{itemize}
\item We propose ConstStyle, a novel domain generalization framework that projects all data into a unified domain, addressing both training and testing phases to improve generalization on unseen domains.
%    \item We present \textit{ConstStyle}, a novel domain generalization approach that simultaneously addresses both the training and testing processes, aiming at boosting model's performance on both seen and unseen domains. The core idea of ConstStyle is to employ a unified target domain onto which all data is projected prior to both training and testing.
 %   In this way, our method enhances the model's robustness in handling unseen domains, even when only a limited number of seen domains are available. Our comprehensive experiments across various scenarios with several benchmark datasets, including PACS, Digit5, and Duke-Market101 show that ConstStyle improve the accuracy by ...\lecomment{\% on seen domains and ...\% on unseen domains}.
   % \item We propose an effective algorithm for identifying such a unified domain. %based on data from the seen domains, with the goal of optimizing the model’s performance on these domains during training. 
%    The algorithm is underpinned by a theoretical analysis, providing a solid foundation for its reliability and performance. 
    \item We present a theoretically grounded algorithm for defining such the unified domain, ensuring its reliability and effectiveness in reducing domain shifts.
    \item We introduce an alignment algorithm that projects unseen samples onto the unified domain during testing, with the goal of reducing information loss and closing the domain gap between the testing and training domains. This algorithm is backed by a theoretical analysis that establishes performance bounds for the model on unseen domains.
   % \item Extensive experiments on multiple benchmarks demonstrate that ConstStyle significantly outperforms existing methods, achieving up to 19.82\% improvement in accuracy. 
   \item Extensive experiments in various scenarios with benchmark datasets: PACS, Digit5, and Duke-Market101, show that ConstStyle improves precision up to 19.82\%.

\end{itemize}
\section{Related works}

% Domain Generalization (DG) has become a pivotal research area focused on enhancing the robustness and reliability of models, particularly in high-stakes applications. DG aims to achieve out-of-distribution generalization, allowing models to perform effectively on unseen domains by utilizing one or more source domains during training.
% To tackle the challenges associated with domain shifts, a range of methodologies have been proposed, which can be broadly categorized into the following approaches:

Domain Generalization (DG) is a pivotal research area aimed at enhancing model robustness and reliability, especially in high-stakes applications.
DG focuses on out-of-distribution generalization, enabling models to perform well on unseen domains by training on one or more source domains. Various methods have been proposed to address domain shifts, generally categorized into the following approaches.

\noindent \textbf{Invariant Representation Learning.} Invariant representation learning seeks to extract features consistent across domains. Domain alignment methods, such as CIDDG \cite{li2018deep}, minimize distributional differences across domains. 
Rather than enforcing invariance across all features, disentangled feature learning approaches \cite{chattopadhyay2020learning, piratla2020efficient} separate features into domain-specific and domain-invariant components, then learn them simultaneously. 
To further enhance this, authors in \cite{chen2023domain} introduce RIDG, a method that learns to ensure representations for samples within the same class remain consistent across domains by utilizing a rationale matrix and rationale invariant loss function, fostering improved generalization.
Additionally, normalization techniques \cite{pan2018two, choi2021robustnet} remove style information to produce invariant representations. 
Despite their promising results, these methods require a large number of domains to effectively extract invariant features, posing challenges when deploying the model in real-world environments.

\noindent \textbf{Data Augmentation.} %Data augmentation is a powerful technique to expand the training dataset and improve model generalization. 
Numerous strategies, based on image level, such as AugMix \cite{hendrycks*2020augmix} and CutMix \cite{yun2019cutmix} are developed to achieve robust augmentation. 
Mixup \cite{zhang2018mixup} goes further by using pairwise linear interpolation in both image and label spaces, while Manifold Mixup \cite{verma2019manifold} extends this to the feature level, boosting generalization. 
Additionally, \cite{volpi2018generalizing} enhances the robustness by generating adversarial examples from hypothetical target domains, thereby strengthening the robustness of the model. 
Taking a different approach based on the observation that style statistics (mean and covariance) capture essential style information specific to each domain, style augmentation increases the training data quantity. 
For instance, StyDeSty \cite{liu2024stydesty} uses the stylization module to generate various style versions given a source domain. TF-Cal \cite{zhao2022test} utilizes linear combinations of two seen styles and combinations of a seen style with the representative style, while MixStyle \cite{zhou2021domain} mixes the seen style within batches to increase the diversity of the source domain.
DSU \cite{li2022uncertainty} further estimates feature statistic uncertainty to sample new style features, simulating out-of-distribution domains, while CSU \cite{zhang2024domain} incorporates feature correlation in style mixing to retain semantic consistency, and Style Neophile \cite{kang2022style} selects style prototypes from a style queue based on Maximum Mean Discrepancy, capturing source style distributions.
However, as shown in our analysis \ref{sec:affect}, training with numerous domains does not always enhance model performance due to the dissimilarity between source and unseen domains.

To address the aforementioned challenges, we introduce ConstStyle, a novel approach that alleviates the domain shift problem by projecting all domains into a unified space. This reduces the impact of domain limitations while enabling the model to learn consistent features, thereby enhancing its generalizability.

% \noindent \textbf{Invariant Representation Learning.}
% Invariant representation learning aims to extract features that remain consistent across different domains. 
% Domain alignment methods \cite{li2018deep} achieve this by minimizing the differences between various distributions. 
% Conversely, disentangled feature learning approaches \cite{chattopadhyay2020learning}; \cite{piratla2020efficient} decompose features into domain-specific and domain-invariant components and learn them concurrently. 
% Additionally, normalization-based techniques \cite{pan2018two}; \cite{choi2021robustnet} can also be employed to eliminate style information, thereby producing invariant representations. 
% \cite{chen2023domain} utilized a rationale matrix with the rationale invariant loss to promote the consistency of representations for samples within the same class, ensuring that their representations are identical. 

% However, the aforementioned methods requires subtaintial number of source domains to yield the optimal performance, which hinder the deployment in realworld since the number of source domains is limited. 
% To this end, we introduce a novel approach that addresses the domain shift problem by projecting all domains into a unified domain to mitigate the impact of limited domains, thus increasing the generalibility of the model. 
% This unified projection not only simplifies and enhances the learning process but also ensures more efficient and robust model performance, thereby improving the model’s ability to generalize effectively to arbitrary unseen domains.

\section{Our Proposed ConstStyle}
\label{sec:proposal}
\begin{figure}[t]
    \centering
    \includegraphics[width=1\linewidth]{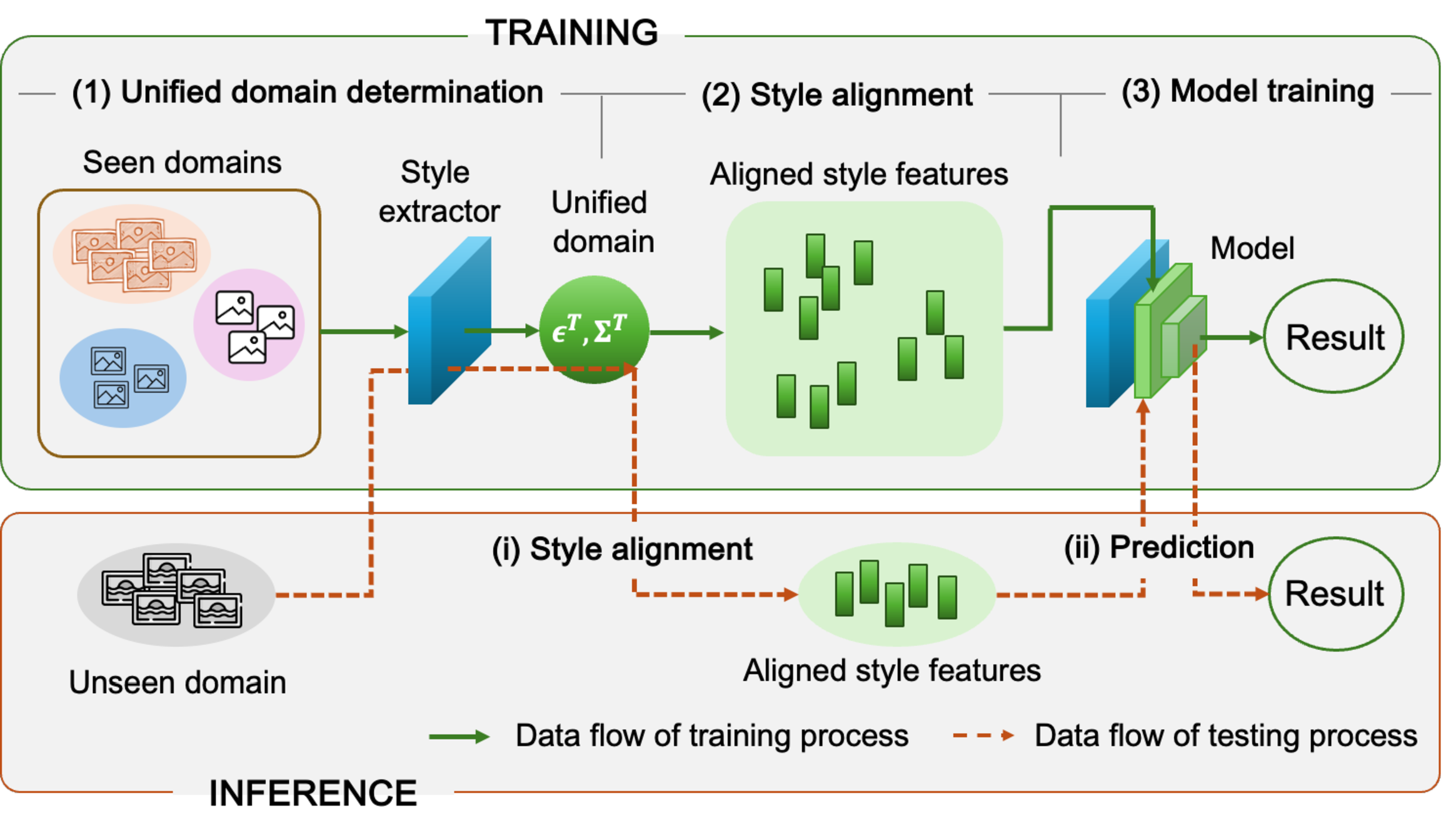}
    \caption{Overview of ConstStyle.} 
    \label{fig:overview}
    \vspace{-15pt}
\end{figure}
\subsection{Preliminaries}
\noindent \textbf{Notations and Definitions.} 
Throughout this paper, we assume that there are $N$ seen domains $\mathcal{S}_1, ..., \mathcal{S}_N$ and $M$ unseen domains $\mathcal{U}_1, ..., \mathcal{U}_M$ (where $N$ and $M$ are not known in advance). 
We focus on the classification task.
Let $\omega$ denote the model of interest, which consists of two components: a representation learning module, denoted by $\theta$, and a classifier head, denoted by $\zeta$.  
As noted in previous studies \cite{zhou2021domain}, the intermediate layers of $\theta$ often capture domain-specific style information. Thus, $\theta$ can be decomposed into two parts: $\theta = \theta_f(\theta_s(x))$, where $\theta_s$ serves as the style extractor, and $\theta_f$ generates the final representation of the image \cite{huang2017adain}. 
\vspace{-1pt}
\begin{definition}[Instance Style] 
Given an input image $x$, let $z_x \in \mathbb{R}^{C \times H \times W}$ be the style feature of $x$, i.e., $z_x = \theta_s(x)$.
We define by $\epsilon_x$ the style statistic of $x$, which captures the channel-wise mean and variance of $z_x$. Specifically, we express $\epsilon_x$ as $\epsilon_x = concat(\mu_x, \sigma_x)$, where $\mu_x \in \mathbb{R}^C$ and $\sigma_x \in \mathbb{R}^C$ are defined as follows:
\begin{align}
\nonumber
    \mu_{x_c} &= \frac{1}{HW}\sum_{h=1}^H \sum_{w=1}^W z_{x_{c, h, w}}, \\
\nonumber    \sigma_{x_c} &= \sqrt{\frac{1}{HW}\sum_{h=1}^H \sum_{w=1}^W (z_{x_{c, h,w}}-\mu_{x_c})^2}.
\end{align}
\end{definition}
\begin{definition}[Domain Style]
    Let $\mathcal{S}$ be a domain and $D_\mathcal{S}$ a set of data samples belonging to $\mathcal{S}$, $D_\mathcal{S} = \{(x_1, y_1), ..., (x_{|D_\mathcal{S}|}, y_{|D_\mathcal{S}|})\}$, and let $\epsilon_{x_i}$ be the style statistic of instance $x_i$. 
    We define the style of domain $D_\mathcal{S}$, denoted as $\mathcal{P}_\mathcal{S}$ as multivariate normal distribution representing the style statistics of $D_\mathcal{S}$'s elements, i,e,., $\{\epsilon_{x_i}\}_{i=1}^{|D_{\mathcal{S}}|}$. The mean $\epsilon_\mathcal{S}$ and variance $\Sigma_\mathcal{S}$ of $\mathcal{P}_\mathcal{S}$ is calculated as follows: 
    \begin{align*}
        \epsilon_{\mathcal{S}} = \frac{1}{|D_\mathcal{S}|} \sum_{i=1}^{|D_\mathcal{S}|} \epsilon_{x_i},\Sigma_\mathcal{S} = \frac{1}{|D_\mathcal{S}|}\sum _{i=1}^{|D_\mathcal{S}|} (\epsilon_{x_i} - \epsilon_{\mathcal{S}})^T (\epsilon_{x_i} - \epsilon_{\mathcal{S}}).
    \end{align*}
\end{definition}

\noindent \textbf{Problem Formulation.}
Let $(\mathcal{X}, \mathcal{Y})$ denote the space of inputs and labels.
The DG problem asks us to identify a model $\omega^*$, which is trained on all seen domains $\{\mathcal{S}_k\}_{k=1}^N$, and performs well on both seen domains $\{\mathcal{S}_k\}_{k=1}^N$ and unseen domains $\{\mathcal{U}_j\}_{j=1}^M$ ($j = 1, ... M$).
This can be formulated as follows: 
\begin{multline}
\nonumber
    \omega^* = \arg \min_{\omega} \left ( \sum_{j=1}^M \mathop{\mathbb{E}}_{(u,y) \in \mathcal{U}_j}\left[ l(\omega(u), y) \right]  \right. \\
    \left. + \sum_{k=1}^N \mathop{\mathbb{E}}_{(x,y) \in \mathcal{S}_k}\left[ l(\omega(x), y) \right] \right ),
    \label{formulation2}
\end{multline}
% where $l$ is the loss function (typically the cross entropy loss).
where $l$ is the loss function, which is varied depending on the task (e.g. cross entropy loss for the classification task).
% , defined as: $l(\omega(x),y) = -\sum_{c=1}^C y_c* \log(p_c)$, with $y_c$ represents one-hot encoded vector of class label $c$ and $p_c$ represents predicted probability for each class $c$, obtained by applying softmax on the model output $\omega(x)$.

\subsection{Overview}
Figure \ref{fig:overview} presents the workflow of ConstStyle, designed to enhance accuracy across both seen and unseen domains by addressing both training and testing phases. In the training phase, ConstStyle follows three main steps: (i) determining the style of the unified domain; (ii) transforming the style of all samples in the training data set (from the seen domains) to match the unified domain’s style; and (iii) training the model using these style-aligned samples.
In the testing phase, samples from the unseen domain are first adjusted to align with the unified domain style before being processed by the trained model for inference.

A central challenge is defining an appropriate unified domain, especially since unseen domains are unknown during training. To tackle this, we introduce an algorithm that constructs the unified domain using only information from seen domains, optimizing its selection for better performance on seen data.
However, since the unified domain is defined solely using information from seen domains, a significant gap may exist between the unified and unseen domains. This can lead to substantial data distortion when mapping unseen samples to the unified domain, potentially decreasing model performance.
To address this issue, we propose a partial alignment algorithm that efficiently aligns the test samples to the unified domain. This solution balances the alignment of the testing sample style with the trained style (thus addressing the domain shift) while preserving the essential information of the testing data (thus avoiding performance degradation due to excessive data distortion).

%\sout{In the training phase, ConstStyle follows three key steps: first, identifying the style of the target domain; second, transforming the style of all samples in the training data set (from the seen domains) to match the target domain’s style; and third, training the model using these transformed samples by style.  During the testing phase, samples from the unseen domain are first adjusted to match the target domain style before being fed through the trained model for inference.} \mt{Is this a common practice, adjusted the testing to match our domain's style (which was trained on) and test? Due to adjusting the style, will the inference accuracy increase?} \solved
The details of the algorithm for determining the style of the unified domain are presented in Section \ref{subsec:determination}, while the algorithms for the training and testing process are detailed in Sections \ref{subsec: training} and \ref{subsec:inference}.

\subsection{Unified Domain Determination}
\label{subsec:determination}

The unified domain is designed to maximize the model’s performance on seen domains. To this end, our unified domain is defined as follows.
%\sout{Our algorithm is based on the following theorem:}

\begin{definition}[Unified Domain]
\label{def:unified domain}
    Given $N$ seen domains $\{S_k\}_{k=1}^N$, each $S_k$ has an associated domain style $\mathcal{P}_{S_k}$ follows a distribution $\mathcal{N}(\epsilon_{S_k}, \Sigma_{S_k})$. The unified domain $\mathcal{T}$ is the one with the domain style that serves as the Barycenter of $\{\mathcal{P}_{S_k}\}_{k=1}^{N}$ (denoted as  $\mathcal{B}$). 
    Specifically, the Barycenter \cite{alvarez2016fixed}  is a distribution, i.e., $\mathcal{B} \sim \mathcal{N}(\epsilon_B, \Sigma_B)$, that intuitively minimizes the total distance to $\{\mathcal{P}_k\}_{k=1}^{N}$, where  $\epsilon_B$ is determined as follows:
   \begin{equation}
   \nonumber
        \epsilon_B =  \frac{1}{N} \sum_{k=1}^N \epsilon_{S_k}. 
   \end{equation}
The covariance matrix $\Sigma_B$ is obtained by solving an iterative optimization problem. Starting with an initial covariance matrix $\Sigma_{B_0} = \frac{1}{N} \sum_{k=1}^N \Sigma_{S_k}$, the update formula at each iteration $i$ is given by:
\begin{equation}
\nonumber
       \Sigma_{B_{i + 1}} =  \frac{1}{N} \sum_{k=1}^N \left ( \Sigma_{B_i}^{\frac{1}{2}} \Sigma_{S_k} \Sigma_{B_i}^{\frac{1}{2}} \right )^{\frac{1}{2}}.
\end{equation}     
\end{definition}

Next, we explain the rationale behind our unified domain and present a theoretical analysis of the model's performance when trained with data projected onto this domain.

\noindent \textbf{Theoretical Analysis.}
Let $D_k$ denote the dataset from domain $S_k$, and $D^T_k$ represent $D_k$ after it has been mapped to the unified domain. Naturally, the model $\omega$ often achieves optimal performance on the seen domain $S_k$ when trained directly on the original dataset $D_k$.
Therefore, we aim at designing a unified domain such that training $\omega$ on $D^T_k$ achieves performance comparable to training on $D_k$. 
\begin{lemma}
    Let $\omega^*$ and $\omega^T$ be the models trained using $D_k$ and $D^T_k$, respectively. 
    Let us denote by $L^{\mathcal{S}_k}$ and $L^{\mathcal{S}_k^T}$ the empirical losses of $\omega^*$ and $\omega^T$ calculated over $D_k$ and $D^T_k$, respectively. 
    Then, the gap between $L^{\mathcal{S}_k}$ and $L^{\mathcal{S}_k^{\mathcal{T}}}$ is bounded as follows: 
\begin{equation}
    L^{\mathcal{S}_k^{T}} - L^{\mathcal{S}_k} \leq \beta \times \left (\mathcal{D}_\mu(\mathcal{T}, \mathcal{S}_k) +\mathcal{D}_\sigma(\mathcal{T},\mathcal{S}_k) \right ),
\end{equation}
where $\mathcal{D}_\mu(\mathcal{T}, \mathcal{S}_k)$ and $\mathcal{D}_\sigma(\mathcal{T},\mathcal{S}_k)$ denote the distances between the means and standard deviations of distributions of unified domain $\mathcal{T}$ and seen domain $\mathcal{S}_k$, respectively; $\beta$ is the upper bound of Lipschitz coefficient of the loss function on all seen domains. (The proof is shown in Suppl. B.1).
% and $\beta_\mu$, $\beta_\sigma$ are mean and standard covariance coefficient values of ...  model tested on seen domains by ERM.
\end{lemma}

\begin{theorem}\label{thm:1}
The disparity in empirical losses computed across all the seen data, between the model trained on data projected onto the unified domain and those trained on the original data from the seen domains \(\{S_k\}_{k=1}^{N}\), is bound by the following inequality:
\begin{equation}
\label{eq:distance optim}
        \sum_{k=1}^N \left ( L^{\mathcal{S}_k^{T}} - L^{\mathcal{S}_k} \right ) \leq \beta \times \sum_{k=1}^N \left (   \mathcal{D}_\mu(\mathcal{T}, \mathcal{S}_k) 
        +\mathcal{D}_\sigma(\mathcal{T},\mathcal{S}_k) \right ). 
    \end{equation}
(The proof is presented in Suppl. B.2)
% \mt{\bf(Proof in Appendix...)}
\end{theorem}
Theorem \ref{thm:1} suggests that the unified domain should be selected to minimize the distance on the right side of (\ref{eq:distance optim}). %This can be accomplished by adopting the Barycenter.

\noindent \textbf{Practical Algorithm.}
Now, assuming that we have $n$ training samples: $\{(x_i, y_i)\}_{i=1}^{n}$, with each $x_i$ associated with a style statistic $\epsilon_{x_i}$. If the seen domains $\{S_k\}_{k=1}^{N}$ related to these samples are known, Definition \ref{def:unified domain} could be applied directly to determine the unified domain style.

In practical scenarios, however, we usually have only the training samples without specific domain labels. To address this, we introduce an efficient algorithm to estimate the style statistics of the seen domains before applying Definition \ref{def:unified domain}. Specifically, we utilize a Gaussian Mixture Model (GMM) to capture the distribution of the style statistics $\{\epsilon_{x_i}\}_{i=1}^{n}$. The GMM Expectation-Maximization Algorithm is then employed to cluster $\{\epsilon_{x_i}\}_{i=1}^{n}$ into $N'$ distinct groups. The number of clusters $N'$ is treated as a hyperparameter (which can be set to the number of seen domains if this information is known). Since samples from the same domain typically exhibit similar style statistics, each cluster approximately represents style statistics from the same domain. Consequently, the distribution of each cluster can be considered as an approximate style statistic for that domain. Finally, Definition \ref{def:unified domain} is applied to the normal distributions associated with these clusters to establish the unified domain's style statistic.

In practice, calculating the exact Barycenter can be computationally intensive. Thus, a straightforward yet effective approximation involves defining domain style of the unified domain, i.e., $\mathcal{P}_T = \mathcal{N}(\epsilon^{T}, \Sigma^{T})$, by averaging the style distribution of the clusters as follows:
\begin{align}
    \epsilon^{T} = \frac{1}{N'} \sum _{k=1}^{N'} \epsilon_{\mathcal{C}_k}, 
    \Sigma^{T} = \frac{1}{N'} \sum _{k=1}^{N'} \Sigma_{\mathcal{C}_k},
\end{align}
where $N'$ is the number of clusters, and $\epsilon_{\mathcal{C}_k}$, $\Sigma_{\mathcal{C}_k}$ are mean and covariance matrix of the Gaussian distribution associated with cluster $\mathcal{C}_k$, respectively. 
% \ndg{The domain style is updated periodically throughout training to incrementally refine its values, aiming to find the optimal unified domain representation.}. 
Details of the unified domain determination is provided in Algorithm 2 (Supplementary).
\subsection{Training Process}
\label{subsec: training}
%\sout{The training process consists of two main stages: \textit{Initial training} and \textit{Style-transformation training}}. \mt{What is the goal of initial training, what is the goal of style-transformation training.. Give a bit higher overview before describing each in details} \solved

Style transformation involves adapting the original style of training samples to match the style of the unified domain. This process relies on a trained model to extract the style features from the samples. 
To achieve this, we split the ConstStyle training process into two stages: \textit{Initial Training} and \textit{Unified-Style Training}. In the initial training phase, we train a model using the original training data to develop a feature extractor capable of capturing the style features of the samples. In the subsequent phase, the feature extractor is used to transform all training data to match the style of the unified domain. The model is then trained using these style-aligned samples.
{To save training costs, we perform the initial training phase for only a few epochs (instead of training until convergence), to establish the initial unified domain. Afterward, we utilize the feature extractor from unified-style training.}

\noindent{\bf Initial Training.} Initially, the model $\omega$ is trained on the original training dataset using the traditional Empirical Risk Minimization (ERM) approach. Assume that the model obtained after the first training epoch is $\omega^o = \zeta^o(\theta_f^o(\theta_s^o))$, where $\theta^o_s$ denotes the style feature extractor, $\theta^o_f$ the remaining encoder component, and $\zeta^o$ the classification head.
Then, each training sample $x_i$ is fed through $\theta^o_s(.)$ to get its original style feature $z^o_{x_i} = \theta_s^o(x_i)$.
These style features are used to determine the {initial} unified domain $\mathcal{T}^o$ as described in Section \ref{subsec:determination}. 
Once the unified domain is acquired, the training process shifts to the second phase, where the model is trained with data transformed to the unified domain through a style transformation procedure.

\noindent {\bf Unified-Style Training.}
{$\omega^o$ and $\mathcal{T}^o$ are used as the initial points for this training process}. 
At each training epoch during this phase, each image $x_i$ in the training batch is fed through {the current style extractor} $\theta_s(.)$ to get its original style feature $z_{x_i} = \theta_s(x_i)$. This original style statistic is then aligned with a random style statistic $\epsilon_s = (\mu_s, \sigma_s)$ sampled from the unified domain's style distribution $\mathcal{P}_\mathcal{T} = \mathcal{N}(\epsilon_T, \Sigma_T)$ as follows:
\vspace{-5pt}
\begin{equation}
    z^T_{x_i} = \sigma_s \times \frac{z_{x_i}-\mu_x}{\sigma_x} + \mu_s.
    \vspace{-5pt}
\end{equation}
After transformation, the new style feature $z^T_{x_i}$ is fed into the model $\omega$ for further training, i.e., $\omega = \zeta(\theta_f(z^T_{x_i}))$.
{Back-propagation is performed across all components of $\omega$, including the feature extractor $\theta_s(.)$. The unified domain $\mathcal{T}$ is periodically updated every $\gamma$ epochs, a strategy that improves unified domain quality while preserving training stability.} It is worth noting that, through this projection process, the model is trained on $E*D$ different variations of style features (where $E$ is the number of training epochs and $D$ is the total seen data), all aligned with the unified domain. 
This strategy not only expands the training dataset but also enriches the diversity of style features specific to the unified domain, thereby enhancing the model's adaptability to it. Details of the training process are presented in Algorithm 1 in Supplementary.
\subsection{Inference Process}
\label{subsec:inference}
To bridge the domain gap between test and training data, we introduce a novel approach that aligns the style of test samples with the unified domain prior to inference. The challenge here is that since the unified domain is entirely defined based on the seen domains, the gap between the unified domain and the test domains (unseen domains) can be quite large, leading to the potential loss of original characteristics in the data after being aligned with the unified domain\footnote{This issue does not arise with data from seen domains, as the unified domain is designed to closely align with them.}. To address this issue, we employ a partial projection strategy that balances transforming the style of test samples to match the unified domain while preserving their original characteristics.
%our approach is to use partial projection, or in other words, to strike a balance between transforming the style of the test sample to match the unified domain’s style and preserving the sample’s original features.%\mt{Again, is it a typical and common approach in testing unseen domains?} \solved

Specifically, let $\omega^* = \zeta^*(\theta_f^*(\theta_s^o))$ be the model obtained after the training phase. 
In the inference phase, each testing sample $u$ is firstly fed into $\theta_s^o$ to get the style feature $z^o_{u}$.
This style feature is then partially aligned with the unified domain's style to generate a new style feature, denoted as $z^T_{u}$. This $z^T_{u}$ is subsequently input into $\zeta^*(\theta_f^*(.))$ to yield the final prediction result.
The formula below is used to align the style $z^o_{u}$ of a test sample $u$:
\begin{multline}
    z^T_{u} = \left ( \alpha \times \sigma_u + (1 - \alpha) \times \sigma_T \right ) \frac{z^o_{u} - \mu_u}{\sigma_u} \\
    + \left ( \alpha \times \mu_u + (1 - \alpha) \times \mu_T \right ),
\end{multline}
where $(\mu_u, \sigma_u)$ is the original style statistic of $u$, and $\epsilon_T = (\mu_T, \sigma_T)$ is the mean; $\alpha$ is a hyperparameter in the range of $(0, 1)$ controlling the extent to which the original feature is preserved.
Specifically, $\alpha = 0$ indicates that the testing data is completely mapped to the unified domain, while $\alpha = 1$ keeps the test data in its original state. We perform an ablation study with different values of $\alpha$ to examine its effects and assess our alignment algorithm (see Supplementary E.4).
Additionally, we present below a theorem that establishes a bound on the distance between the empirical losses of the model on unseen and seen domains.
% The rationale behind this partial projection algorithm is that that there may be a substantial gap between the unseen domain and the target domain. This issue does not arise with data from seen domains, as the target domain is designed to closely align with them. Therefore, fully projecting samples from the unseen domain could lead to information loss and may negatively impact predictive performance.
\begin{theorem}
Let $\mathcal{S}$ and $\mathcal{U}$ be the set of data from all seen and unseen domains, respectively; $D_{S^T}$ and $D_{U^T}$ be the set of seen data and unseen data after %either fully or partially 
projected onto the unified domain using our algorithm; $L^{\mathcal{U}^{T}}$ and $L^{\mathcal{S}^{T}}$ be the empirical losses of models trained by $D_{S^T}$ and $D_{U^T}$ calculated over the seen and unseen data, respectively. 
The difference of $L^{\mathcal{U}^{T}}$ and $L^{\mathcal{S}^{T}}$ is bounded by the following inequality: 
    \begin{align}
        L^{\mathcal{U}^{T}} - L^{\mathcal{S}^{T}} & \leq  \alpha \times \beta \times (\mathcal{D}_\mu(\mathcal{U, \mathcal{T}})+\mathcal{D}_\sigma(\mathcal{U}, \mathcal{T})) \\
        \nonumber
        & + \epsilon \times \sqrt{2.Tr(I)},
    \end{align}
    where $\mathcal{D}_\mu(\mathcal{T}, \mathcal{U})$ and $\mathcal{D}_\sigma(\mathcal{T}, \mathcal{U})$ represent the distances between the mean and variance of the distributions of $\mathcal{T}$ and $\mathcal{U}$. $Tr(I)$ represents the trace of the identity matrix $I$, which is the sum of the diagonal elements of the matrix. $I$ is the identity matrix with dimensions $C \times H \times W$, where $C, H, W$ are the channel, height, and width dimensions of the output of $\theta_s$. (The proof is provided in Suppl. B.3).
    %\mt{\bf(Proof in ...)}
    % \lecomment{Additionally, $\mathcal{D}_{max}(\mathcal{N}(0,I))$ denotes the maximum distance between two points within a standard Gaussian distribution.}
\end{theorem}
We provide details of the inference process in Algorithm 3 in Supplementary.
\section{Experimental Evaluation}
\label{sec:experiments}
\begin{table}[t]
    \centering
    \resizebox{\linewidth}{!}{
        \begin{tabular}{l|c|cccc|r}
         \toprule
        \multicolumn{1}{c|}{\multirow{2}{*}{\textbf{Method}}} & \multirow{2}{*}{\textbf{Venue}} & \multicolumn{4}{c|}{\textbf{In-domain combinations}} & \multirow{2}{*}{\textbf{Avg}} \\
        \cmidrule{3-6}
        & & A, C, S & P, C, S & P, A, S & P, A, C \\ \midrule
         ERM & - & 95.02 & 95.91 & 95.75 & 97.08 & 95.94 \\
         MixStyle \cite{zhou2021domain} & ICLR 2021 & 94.66 & 95.78 & 96.01 & 96.59 & 95.76 \\
         DSU \cite{li2022uncertainty} & ICLR 2022 & \nduong{94.78} & \nduong{96.65} & \nam{96.39} & \nam{97.40} & \nduong{96.30} \\
         CSU \cite{zhang2024domain} & WACV 2024 & \nduong{94.78} & 96.52 & 96.13 & 97.07 & 96.12 \\ 
         \midrule
         \textbf{ConstStyle} & \textbf{Ours} & \nam{95.37} & \nam{97.02} & \nam{96.39} & \nduong{97.24} & \nam{96.50} \\
         \bottomrule
        \end{tabular}
    }
    \caption{In-domain performance of ConstStyle compared with baselines on PACS datasets. P, A, C, S denote Photo, Art, Cartoon, Sketch. The best result is colored \nam{purple} and the second best result is colored \nduong{blue}.}
    \label{tab:pacs-in-domain}
    \vspace{-10pt}
\end{table}
We conducted a series of experiments to assess ConstStyle's effectiveness across various scenarios. 
%These experiments were designed to examine the model's ability to generalize, maintain stability, and adapt to distributional shifts. 
The evaluation covered three primary tasks: %, including 
image classification, image corruption, and instance retrieval, offering a thorough analysis of the method's robustness under different conditions.

\subsection{Settings} 

\noindent \textbf{Image Classification.} We address the style-shift problem and evaluate our method on the PACS dataset \cite{li2017deeper}, a Domain Generalization benchmark with four styles (Photo, Art, Cartoon, Sketch). Following \cite{Yu_2024_CVPR}, we conduct experiments under two scenarios: (1) a single unseen domain, where the model trains on three domains and tests on the fourth \cite{zhou2021domain, li2022uncertainty, zhang2024domain}, and (2) multiple unseen domains, where training is further restricted to assess generalization. Additionally, we evaluate the Digits5 dataset \cite{yann1998gradient} to improve robustness across five domains.
% We focus on style-shift problem and evaluate our proposed method on the PACS dataset \cite{li2017deeper}, a standard Domain Generalization benchmark with four styles: Photo, Art, Cartoon, and Sketch, using a ResNet18 backbone pretrained on ImageNet \cite{zhou2021domain, li2022uncertainty}. 
% Following recommendations in \cite{Yu_2024_CVPR}, we perform extensive experiments under two scenarios: (1) a single unseen domain, where the model trains on three domains and tests on the fourth, as in \cite{zhou2021domain, li2022uncertainty, zhang2024domain} setups; and (2) multiple unseen domains, where training is further restricted, testing the model’s ability to generalize across diverse domains. 
% Furthermore, we evaluated the Digits5 dataset \cite{yann1998gradient} to enhance robustness across five domains.

\noindent \textbf{Image Corruption.} We further assess the robustness of our method against image corruption using the CIFAR10-C dataset \cite{hendrycks2018benchmarking}, which includes 19 types of corruption at five severity levels. 
Higher levels indicate stronger corruption. CIFAR10 serves as the source domain and CIFAR10-C as the target domain.

\noindent \textbf{Instance Retrieval.} For the instance retrieval task, we evaluate re-ID methods, matching individuals across camera views. Market1501 \cite{zheng2015scalable} and Duke \cite{ristani2016performance} are used interchangeably for training and testing. 
Performance is assessed using the ranking accuracy and mean average precision (mAP).
In all experiments, we set $\alpha=0.6$ for PACS dataset and $\alpha=0.5$ for Digit5 dataset and report the results with ERM refers to the approach that trains the model using Empirical Risk Minimization loss. Details of experimental settings are provided in Suppl. C.
% In all the results reported below, ERM refers to the approach that trains the model on all seen data using the Empirical Risk Minimization loss.
\begin{table}[t]
    \centering
    \resizebox{\linewidth}{!}{
        \begin{tabular}{l|c|cccc|r}
         \toprule
        \multicolumn{1}{c|}{\multirow{2}{*}{\textbf{Method}}} & \multirow{2}{*}{\textbf{Venue}} & \multicolumn{4}{c|}{\textbf{Domains}} & \multirow{2}{*}{\textbf{Avg}} \\
        \cmidrule{3-6}
        & & Art & Cartoon & Photo & Sketch \\ \midrule
         ERM & - & 77.10 & 77.77 & 96.40 & 68.17 & 79.86 \\
         Crossgrad \cite{shiv2018crossgrad} & ICLR 2018 & 78.12 & 77.90 & 96.64 & 70.64 & 80.82 \\
         Mixup \cite{zhang2018mixup} & ICLR 2018 & 78.71 & 74.53 & 96.16 & 66.24 & 78.91 \\
         Cutmix \cite{yun2019cutmix} & ICCV 2019 & 77.49 & 73.33 & 96.34 & 69.80 & 79.24 \\
         EDFMix \cite{zhang2021exact} & CVPR 2022 & 83.05 & 81.05 & 96.64 & 76.50 & 84.31 \\
         RIDG \cite{chen2023domain} & ICCV 2023 & 80.17 & 78.32 & \nduong{96.82} & 72.32 & 81.90 \\
         MixStyle \cite{zhou2021domain} & ICLR 2021 & 81.25 & 80.03 & \nduong{96.82} & 72.17 & 82.57 \\
         DSU \cite{li2022uncertainty} & ICLR 2022 & 83.94 & 81.10 & 96.23 & 79.05 & 85.08 \\
         CSU \cite{zhang2024domain} & WACV 2024 & \nduong{84.62} & \nduong{82.21} & 96.41 & \nduong{78.11} & \nduong{85.33} \\ 
         \midrule
         \textbf{ConstStyle} & \textbf{Ours} & \nam{85.45} & \nam{82.42} & \nam{96.89} & \nam{82,32} & \nam{86,77} \\
         \bottomrule
        \end{tabular}
    }
    \caption{Performance of ConstStyle compared with baselines on PACS datasets.The best result is colored \nam{purple} and the second best result is colored \nduong{blue}.}
    \label{tab:pacs1}
    \vspace{-10pt}
\end{table}
% \noindent \textbf{Instance Retrieval.} For instance retrieval task, methods are evaluated in person re-identification (re-ID), which aims to match individuals across various camera views. We use two well-known re-ID datasets, Market1501 \cite{zheng2015scalable} and Duke \cite{ristani2016performance}, with one serving for training and the other for testing and comparison. 
% The performance is measured using ranking accuracy and mean average precision (mAP), both expressed as percentages, with the ResNet50 architecture employed for the analysis.

% We compare ConstStyle with nine benchmarks, including Crossgrad \cite{shiv2018crossgrad}, Mixup \cite{zhang2018mixup}, CutMix \cite{yun2019cutmix}, EDFMix \cite{zhang2021exact}, RIDG \cite{chen2023domain}, MixStyle \cite{zhou2021domain}, DSU \cite{li2022uncertainty} and CSU \cite{zhang2024domain}.

\subsection{Image Classification}
\subsubsection{In-domain Performance}
We first evaluate ConstStyle's performance on the seen domains. As shown in Table \ref{tab:pacs-in-domain}, ConstStyle outperforms existing methods in three out of four cases and achieves the highest overall accuracy. Compared to the standard ERM approach, ConstStyle improves performance by 0.56\% and further exceeds the second-best method by an additional 0.2\% in accuracy.

% We first validate our analysis from Section \ref{subsec:determination} through empirical experiments. 
% Specifically, we evaluate our method using an in-domain validation dataset on PACS across various domain combinations.
% In-domain performance is computed as the average accuracy across all validation datasets. 

\subsubsection{Generalization on Multi-domain Classification}
% The comparison of ConstStyle and existing approaches on the image classification task is summarized in Tables \ref{tab:pacs1} and \ref{tab:digit1}.  
% As shown, ConstStyle outperforms all existing methods on PACS dataset and achieves the best performance on four over five unseen domain with respect to Digits5 dataset.
% Concerning the average accuracy over all unseen domains, ConstStyle achieves the best performance on both the two datasets. 
% Specifically, for challenging unseen domains such as Art, Cartoon, and Sketch, ConstStyle improves the performance from $0.45\%$ to $2.27\%$ compared to the second best benchmark. 
% Notably, in the in the Sketch domain, where the style statistics of the test domain diverge substantially from those of the training domains, ConstStyle attains the best improvement gap compared to the other approaches. This finding highlights the effectiveness of ConstStyle in dealing with hard cases when the domain gap is significantly.
% Moreover, regarding Digits5 dataset, ConstStyle continues to demonstrate superior performance over baseline methods, with the most significant improvement gap attained at MNISTM domain, which is again the most challenge one. 
Tables \ref{tab:pacs1} and \ref{tab:digit1} provide a summary of the comparison between ConstStyle and existing approaches on the image classification task. As shown, ConstStyle surpasses all other methods on the PACS dataset and achieves the highest performance on four out of five unseen domains in the Digits5 dataset.
Regarding the average accuracy across unseen domains, ConstStyle leads on both datasets. For particularly challenging domains such as Art, Cartoon, and Sketch, ConstStyle improves performance by $0.21\%$ to $4.21\%$ over the second-best method. In the Sketch domain specifically, where the unseen domain style statistics vary significantly from the training domains, ConstStyle achieves the largest improvement over other methods, underscoring its capability to handle substantial domain gaps. 
On the Digits5 dataset, ConstStyle also shows superior performance over baseline methods, with the greatest improvement observed in the MNISTM domain, which presents the greatest challenge. These results highlight ConstStyle’s strong adaptability and robustness in diverse, difficult settings.

% \begin{table}[t]
%     \centering
%     \resizebox{\linewidth}{!}{
%         \begin{tabular}{l|c|ccccc|r}
%          \hline
%          \multicolumn{1}{|c|}{\multirow{2}{*}{\textbf{Method}}} & \multicolumn{6}{c|}{\textbf{Domains}}& \multirow{2}{*}{\textbf{Avg Acc}}
%          Method & Reference & MNISTM & SVHN & SYN & USPS & MNIST & Avg \\
%          \midrule
%          ERM & - & 67.45 & 77.16 & 86.43 & 96.9 & 98.3 & 85.24 \\
%          Crossgrad & ICLR 2018 & 69.03 & 77.20 & 86.93 & 96.9 & 69.03 & 85.65 \\
%          Mixup & ICLR 2018 & 64.08 & 79.32 & 81.31 & 94.51 & 98.3 & 83.50 \\
%          Cutmix & ICCV 2019 & 65.45 & 79.55 & 84.6 & 95.69 & 98.28 & 84.71 \\
%          EDFMix & CVPR 2022 & 70.95 & 77.55 & 86.94 & 96.88 & 98.3 & 86.15 \\
%          RIDG & ICCV 2023 & 67.35 & 79.18 & 86.86 & \textbf{97.04} & 98.3 & 85.76 \\
%          Mixstyle & ICLR 2021 & 61.48 & 57.18 & 64.17 & 87.58 & 95.15 & 73.11 \\
%          DSU & ICLR 2022 & 67.84 & 77.01 & 87.21 & 96.55 & 98.3 & 85.38 \\
%          CSU & WACV 2024 & 68.25 & 78.6 & 86.7 & 96.4 & 98.1 & 85.61 \\ 
%          ConstStyle & Ours & \textbf{71.51} & \textbf{79.9} & \textbf{87.9} & 96.80 & \textbf{98.3} & \textbf{86.88} \\
%          \bottomrule
%         \end{tabular}
%     }
%     \caption{Performance of ConstStyle compared with baselines on Digits5 datasets.}
%     \label{tab:digit1}
% \end{table}
\begin{table}[t]
    \centering
    \resizebox{\linewidth}{!}{
    \begin{tabular}{l|c|ccccc|c}
    \toprule
    \multicolumn{1}{c|}{\multirow{2}{*}{\textbf{Method}}} & \multirow{2}{*}{\textbf{Venue}} & \multicolumn{5}{c|}{\textbf{Domains}} & \multirow{2}{*}{\textbf{Avg}} \\
    \cmidrule{3-7}
     & & MNISTM & SVHN & SYN & USPS & MNIST & \\ \midrule
    ERM & - & 67.45 & 77.16 & 86.43 & 96.90 & 98.30 & 85.24 \\ 
    Crossgrad\cite{shiv2018crossgrad} & ICLR 2018 & 69.03 & 77.20 & 86.93 & 96.9 & 69.03 & 85.65 \\ 
    Mixup \cite{zhang2018mixup} & ICLR 2018 & 64.08 & 79.32 & 81.34 & 94.51 & 98.30 & 83.50 \\ 
    Cutmix\cite{yun2019cutmix} & ICCV 2019 & 65.45 & \nduong{79.55} & 84.60 & 95.90 & 98.30 & 84.36 \\ 
    EDFMix\cite{zhang2021exact} & CVPR 2022 & \nduong{70.95} & 77.55 & 86.94 & \nduong{96.98} & 98.30 & \nduong{86.14} \\ 
    RIDG\cite{chen2023domain} & ICCV 2023 & 67.35 & 79.18 & 86.86 & \nam{97.04} & 98.30 & 85.95 \\ 
    MixStyle\cite{zhou2021domain} & ICLR 2021 & 61.48 & 57.18 & 64.17 & 87.58 & 96.90 & 73.06 \\ 
    DSU\cite{li2022uncertainty} & ICLR 2022 & 67.84 & 77.01 & \nduong{87.21} & 96.55 & 98.30 & 85.38 \\ 
    CSU\cite{zhang2024domain} & WACV 2024 & 68.25 & 78.60 & 86.70 & 96.40 & \nduong{98.30} & 85.65 \\ 
    \midrule
    \textbf{ConstStyle} & \textbf{Ours} & \nam{71.51} & \nam{79.9} & \nam{87.90} & 96.80 & \nam{98.30} & \nam{86.88} \\ 
    \bottomrule
\end{tabular}
}
\caption{Performance of ConstStyle compared with baselines on Digits5 datasets. The best result is colored \nam{purple} and the second best result is colored \nduong{blue}.}
\label{tab:digit1}
\vspace{-5pt}
\end{table}

\subsubsection{Robustness Against the Numbers of Unseen Domains}
\label{sec:num_unseen_domains}
Previous studies have generally evaluated their methods with limited setups, often using scenarios with only a single unseen domain. To offer a more thorough evaluation, we conduct experiments with various numbers of unseen domains. Specifically, we incrementally increase the number of unseen domains to two for the PACS dataset and three for the Digits5 dataset. Detailed results for the PACS dataset are shown in Table \ref{tab:pacs2}, while those for the Digits5 dataset are provided in the Suppl. D.1.
As shown, ConstStyle continues to achieve the highest accuracy even when the number of seen domains is reduced and the number of unseen domains is increased, demonstrating its strong generalization capability. Specifically, ConstStyle improves accuracy by $1.36\%$ when the test domains are Art and Cartoon, and achieves a notable $5.91\%$ increase over the state of the art when Cartoon and Sketch are the unseen domains. On average, ConstStyle increases overall performance by $2.43\%$, underscoring its consistent effectiveness as the number of training domains varies.
% \vspace{-5pt}
\subsubsection{Impacts of Domain Gap}
\begin{figure}[t]
    \begin{minipage}{\linewidth} 
        \centering
        \includegraphics[width=1\linewidth]{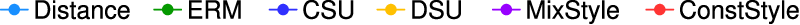}\\
        \vspace{0.1cm}
        \includegraphics[width=0.45\textwidth]{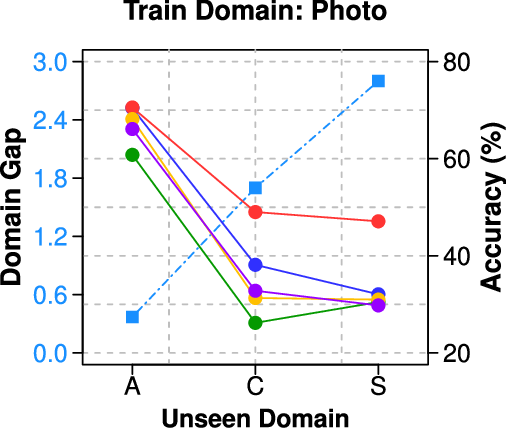}
        \hspace{5pt}
        \includegraphics[width=0.45\textwidth]{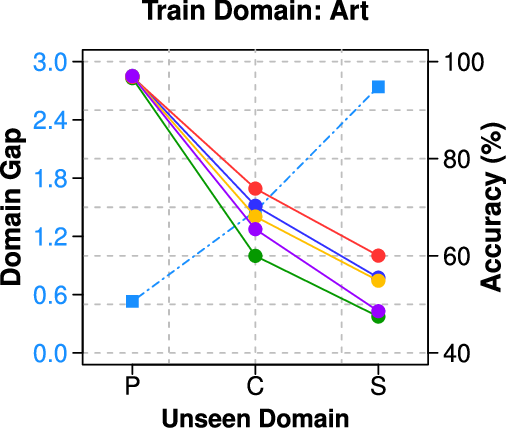}
        \vspace{0.3cm}
        
        \includegraphics[width=0.45\textwidth]{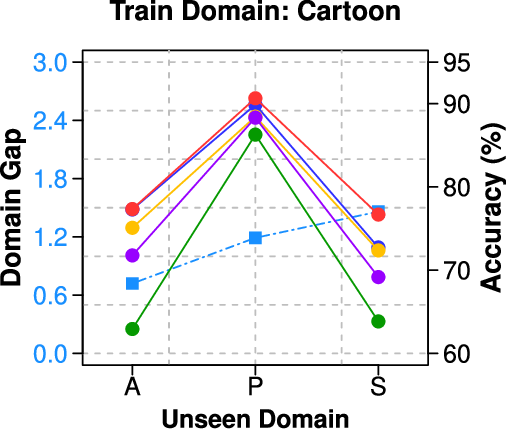}
        \hspace{5pt}
        \includegraphics[width=0.45\textwidth]{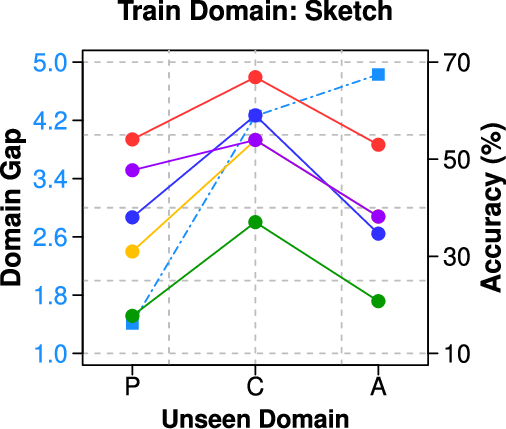}
    \end{minipage}
    \caption{Effect of domain gap. ConstStyle achieves significantly better performance than other methods in handling severe domain gaps.}
    \label{fig:distance_pacs}
    \vspace{-20pt}
\end{figure}
We conduct experiments to evaluate the performance of ConstStyle when confronted with unseen domains that may exhibit varying degrees of distance from the seen domains. Specifically, we train the model on a specific seen domain, then perform inference on various unseen domains, and investigate how the model's accuracy changes. We utilize the Fréchet distance \cite{dowson1982frechet} (also known as the 2-Wasserstein distance) to model the gap between domains' style distributions. 
% as follows
% \begin{equation*}
%     d(P_A, P_B) = |\mu_A - \mu_B|^2 + Tr(\Sigma_A + \Sigma_B - 2(\Sigma_A\Sigma_B))^{1/2}),
% \end{equation*}
% where $\mathcal{P}_A$ and $\mathcal{P}_B$ represent the style distributions of domains $A$ and $B$, respectively, with $\mu_A$, $\mu_B$, and $\Sigma_A$, $\Sigma_B$ denoting their means and covariance matrices. 
The results are shown in Figure \ref{fig:distance_pacs}. In general, as the distance between domains increases, the performance of the methods tends to decrease. However, ConstStyle consistently delivers the highest accuracy and the slowest rate of performance decline across all scenarios. Specifically, when the training domain is Photo, ConstStyle outperforms CSU, with the performance gap ranging from $0.14\%$ to $15.03\%$ as the distance between seen and unseen domains grows. Additionally, ConstStyle shows a performance gap of up to $4.54\%$ when the training domain is Art and $3.95\%$ when the training domain is Cartoon. 
These results highlight the importance of projecting data onto a common domain, which helps mitigate domain gaps and enables the extraction of the most relevant cross-domain features.

\begin{table}[t]
    \centering
    \resizebox{\linewidth}{!}{
        \begin{tabular}{l|c|cccccc|r}
         \toprule
         \textbf{Method} & \textbf{Venue} & A,P & C,P & P,S & A,C & A,S & C,S & \textbf{Avg} \\
         \midrule
         ERM & - & 74.50 & 84.92 & 76.27 & 64.79 & 69.31 & 52.18 & 70.32 \\
         Crossgrad\cite{shiv2018crossgrad} & ICLR 2018 & 74.26 & 85.12 & 76.92 & 64.16 & 69.96 & 51.27 & 70.28 \\ 
         Mixup\cite{zhang2018mixup} & ICLR 2018 & 76.46 & 82.78 & 73.24 & 64.34 & 66.23 & 51.27 & 68.51 \\ 
         Cutmix\cite{yun2019cutmix} & ICCV 2019 & 73.66 & 82.63 & 76.38 & 64.43 & 69.76 & 49.09 & 69.32 \\
         EFDMix\cite{zhang2021exact} & CVPR 2022 & 78.56 & 86.72 & 80.90 & 71.10 & 76.28 & 54.28 & 74.64 \\ 
         RIDG\cite{chen2023domain} & ICCV 2023 & 75.49 & 84.57 & 77.00 & 67.75 & 70.03 & 53.42 & 71.37 \\ 
         MixStyle\cite{zhou2021domain} & ICLR 2021 & 75.60 & 86.79 & 80.54 & 72.10 & 73.34 & 55.58 & 73.99 \\
         DSU\cite{li2022uncertainty} & ICLR 2022 & 80.36 & 86.14 & \nduong{83.56} & 73.49 & \nduong{77.49} & 62.40 & 77.24 \\
         CSU\cite{zhang2024domain} & WACV 2024 & \nduong{82.92} & \nam{87.49} & 83.52 & \nduong{74.59} & 77.42 & \nduong{64.30} & \nduong{78.38} \\ 
         \midrule
         \textbf{ConstStyle} & \textbf{Ours} & \nam{84.64} & \nam{87.49} & \nam{85.28} & \nam{75.95} & \nam{81.29} & \nam{70.21} & \nam{80.81}\\
         \bottomrule
        \end{tabular}
    }
    \caption{Performance comparison on the PACS dataset across six scenarios with different combinations of unseen domains. 
    The best result is colored \nam{purple} and the second best result is colored \nduong{blue}.}
    \label{tab:pacs2}
   \vspace{-5pt}
\end{table}

\subsubsection{Impacts of the Number of Seen Domains}
\label{sec:affect}

\begin{figure}[t]
    \begin{minipage}{\linewidth} 
        \centering
        \includegraphics[width=0.8\linewidth]{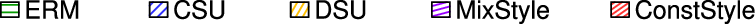}\\
        \vspace{0.1cm}
        \begin{subfigure}[b]{1\linewidth}
            \includegraphics[width=0.44\linewidth]{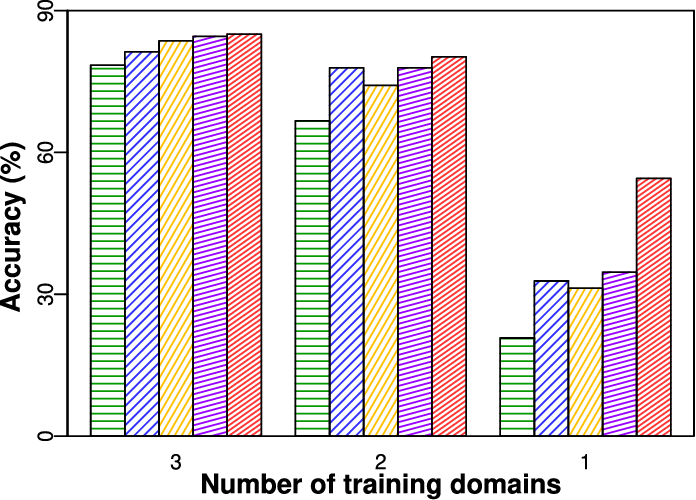}
            \hfill
            \includegraphics[width=0.44\linewidth]{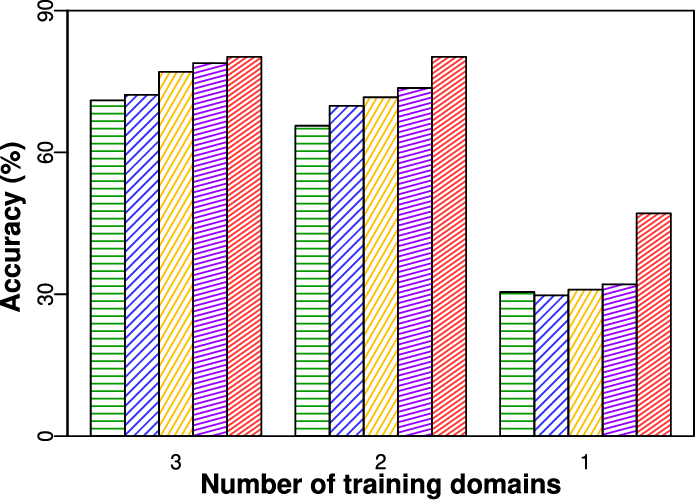}
            \caption{PACS dataset with Test domains are Art (left) and Sketch (right).}
        \end{subfigure}
        \vspace{0.1cm}
        \begin{subfigure}[b]{1\linewidth}
        \centering
            \includegraphics[width=0.44\linewidth]{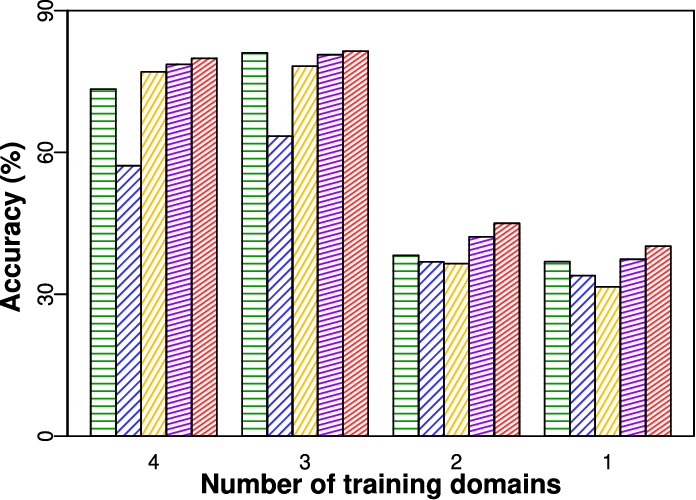}
            \hfill
            \includegraphics[width=0.44\linewidth]{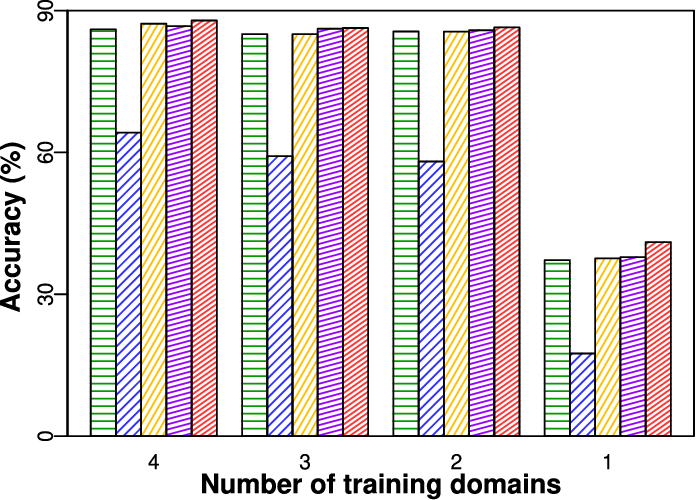}
            \caption{Digit5 dataset with Test domains are SVHN (left) and SYN (right).}
        \end{subfigure}
    \end{minipage}
    \caption{Effects of the number of training domains. ConstStyle consistently delivers the best performance across all scenarios.}
    \label{fig:domain}
    % \vspace{-0.2cm}
\end{figure}

% \begin{figure}[t]
%     \centering
%     \begin{subfigure}[b]{0.31\linewidth}
%         \includegraphics[width=\linewidth]{Figures/insight21.png}
%         \caption{Train domains: A, P\\ Distance: $1.91$ \\Accuracy: $67.44\%$}
%         \label{fig:insight_a}
%     \end{subfigure}
%     \hfill
%     \begin{subfigure}[b]{0.31\linewidth}
%         \includegraphics[width=\linewidth]{Figures/insight22.png}
%         \caption{Train domains: A, C\\ Distance: $2.33$ \\Accuracy: $44.57\%$}
%         \label{fig:insight_b}
%     \end{subfigure}
%     \hfill
%     \begin{subfigure}[b]{0.31\linewidth}
%         \includegraphics[width=\linewidth]{Figures/insight1.png}
%         \caption{Train domains: A\\ Distance: $2.34$\\Accuracy: $46.23\%$}
%         \label{fig:insight_c}
%     \end{subfigure}
%     \caption{Accuracy and distance when train Model with ERM objective with different train domains.}
%     \label{fig:insight}
% \end{figure}
\noindent \textbf{Performance comparison between the models.}
We compare the performance of the methods with the varying number of seen domains (Figure \ref{fig:domain}).
As shown, reducing the number of training domains tends to negatively impact the models' generalizability due to the less diverse features learned from the data. 
However, there are some special cases where training the model with fewer domains is more effective compared with more domains, which will be discussed later. 
Despite the significant decline in performance across methods, ConstStyle exhibits the slowest degradation, which can be observed when comparing with the second-best method, CSU.
Specifically, ConstStyle maintains an accuracy advantage of up to $19.82\%$ when trained on Sketch and tested on Art, and up to $15.02\%$ when trained on Cartoon and tested on Sketch, outperforming CSU. 
In the Digit5 dataset, ConstStyle achieves a performance gap of up to $2.77\%$ when the test domain is SVHN and up to $3.19\%$ when the test domain is SYN, compared to CSU. 
% \begin{figure}[t]
%     \centering
%     \begin{subfigure}[b]{0.31\linewidth}
%         \includegraphics[width=\linewidth]{CVPR/Figures/insight21.jpg}
%         \caption{Train : A, P\\ Distance: $1.91$ \\Accuracy: $67.44\%$}
%         \label{fig:insight_a}
%     \end{subfigure}
%     \hfill
%     \begin{subfigure}[b]{0.31\linewidth}
%         \includegraphics[width=\linewidth]{CVPR/Figures/insight22.jpg}
%         \caption{Train : A, C\\ Distance: $2.33$ \\Accuracy: $44.57\%$}
%         \label{fig:insight_b}
%     \end{subfigure}
%     \hfill
%     \begin{subfigure}[b]{0.31\linewidth}
%         \includegraphics[width=\linewidth]{CVPR/Figures/insight23.jpg}
%         \caption{Train : A\\ Distance: $2.34$\\Accuracy: $46.23\%$}
%         \label{fig:insight_c}
%     \end{subfigure}
%     \caption{Accuracy and distance when train Model with ERM objective with test domain is Sketch.}
%     \label{fig:insight}
% \end{figure}

\begin{figure}[t]
    \centering
    \includegraphics[width=0.9\linewidth]{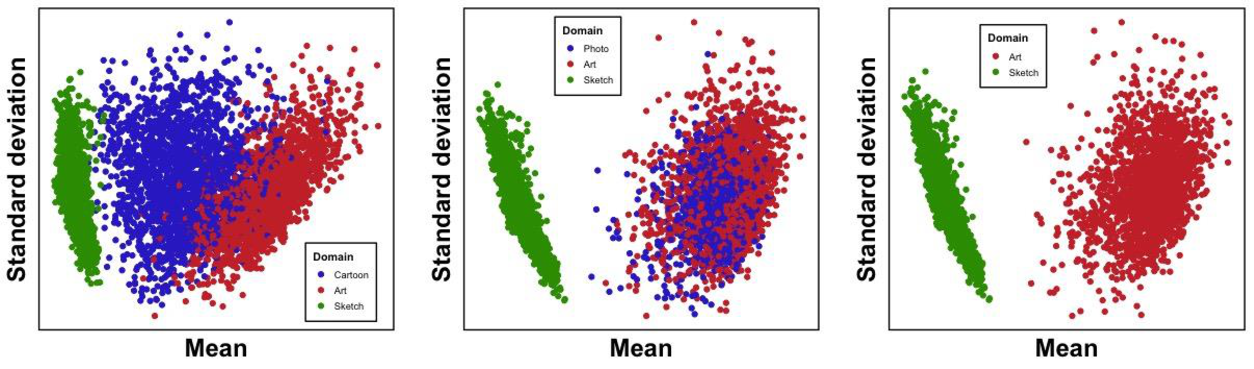}
    % \caption{Visualization of style statistics with colors indicating different domains}
    % \label{fig:insight_a}
    % \hfill
    % \begin{subfigure}[b]{\linewidth}
    %     \includegraphics[width=1.0\linewidth]{Figures/results_class.eps}
    %     \caption{Visualization of feature statistics with colors indicating different classes}
    %     \label{fig:insight_b}
    % \end{subfigure}
    \caption{Style statistics for different domain combinations when training ResNet18 with Sketch as the test domain. Accuracies for each domain combination are 67.44\%, 44.57\%, and 46.23\%.}
    \label{fig:insight}
   \vspace{-10pt}
\end{figure}

\noindent \textbf{When does training with fewer domains result in better performance?} 
As previously discussed, training with a wide range of domains does not necessarily yield better performance. To explore this further, we conducted experiments on PACS dataset, varying the number of training domains. In each experiment, we set Sketch as the test domain and began by training the model with Art as the only seen domain. We then gradually added more domains to the training set, calculating the distance between the seen and unseen domains after each addition.
These results, illustrated in Figure \ref{fig:insight}, reveal that model performance improves only when the added domains decrease the domain gap to the unseen domain.
% For style-shifted datasets, when additional training data successfully narrows this gap, the model’s performance remains stable due to the consistent features and domain attributes. 
% However, if the added domains introduce attributes that act as noise, they can negatively impact the model’s performance as shown in Figure \ref{fig:insight_b} and Figure \ref{fig:insight_c}.

% \vspace{-5pt}
\subsection{Robustness Against Image Corruption}
Table \ref{tab:corrupted} presents the results, revealing that as corruption levels increase, the performance advantage of ConstStyle over existing methods becomes more pronounced. ConstStyle secures the top performance in four out of five cases, with the only exception occurring at the lowest corruption level, where unseen domains closely match the seen domains. On average, ConstStyle achieves the highest performance with a 1.04\% improvement over the second-best method. Compared to the ERM method, which performs best at the lowest corruption level, ConstStyle surpasses it by 0.36\% to 15.83\% across higher corruption levels. Moreover, ConstStyle outperforms other style-based methods with improvements ranging from 0.75\% to 4.96\%, emphasizing its resilience in handling heavily corrupted datasets.
These results underscore ConstStyle's robustness in handling highly corrupted datasets.

% summarizes the results, showing that the performance gap between ConstStyle and existing methods widens as corruption levels increase. ConstStyle achieves the best performance in four out of five scenarios. The only exception occurs at the lowest corruption level, where the unseen domains closely resemble the seen ones. 
% In average, ConstStyle attains the best performance with the improvement gap of 1.04\% compared to the second-best. 
% In comparison to ERM method (which achieves the best performance in level 1), ConstStyle outperforms by 0.36\% to 15.83 in other cases. Additionally, ConstStyle outperforms other style-based methods, with gains from 0.75\% to 4.96\%, underscoring its robustness in handling highly corrupted datasets.

% This results prove the robue
% . It indicates that at lower levels of corruption, specifically at level 1, ERM method attains optimal performance. 
% This observation is consistent with theoretical expectations, as the divergence between the source domain and the target domain is minimal when corruption levels are mild. 
% However, as the corruption severity increases, ConstStyle demonstrates superior performance, consistently outperforming all baseline methods. 

\begin{table}[t]
    \centering
    \resizebox{\linewidth}{!}{
        \begin{tabular}{l|ccccc|r}
        \toprule
         \multicolumn{1}{c|}{\multirow{2}{*}{\textbf{Method}}} & \multicolumn{5}{c|}{\textbf{Level of corruption}} & \multirow{2}{*}{\textbf{Avg}} \\
        \cmidrule{2-6} & level 1 & level 2 & level 3 & level 4 & level 5 & \\ \midrule
         ERM & \nam{87.33} & \nduong{80.29} & 70.57 & 64.81 & 47.80 & 70.16 \\
         MixStyle\cite{zhou2021domain} & \nduong{87.32} & 79.90 & \nduong{71.83}	& \nduong{66.54}	& 58.67 &	\nduong{73.13} \\
         DSU\cite{li2022uncertainty} & 86.91	& 79.40 & 68.68 & 64.14 & 49.44 & 69.71 \\ 
         CSU\cite{zhang2024domain} & 86.73 & 79.87 & 70.66 & 65.51 & \nduong{60.88} & 72.73 \\ 
         \midrule
        \textbf{ConstStyle} & 86.59 & \nam{80.65} & \nam{72.03} & \nam{67.96} & \nam{63.63} & \nam{74.17} \\
         \bottomrule
        \end{tabular}
    }
    \caption{Comparison of methods under different corruption levels, conducted on the CIFAR10-C datasets. The best result is colored \nam{purple} and the second best result is colored \nduong{blue}.}
    \label{tab:corrupted}
\end{table}

\subsection{Generalization in Instance Retrieval}
The results shown in Table \ref{tab:instance_retrieval} highlight that ConstStyle substantially outperforms all other methods in both settings. While other methods encounter difficulties with new domains due to their dependence on generating instances from the existing dataset, ConstStyle addresses this challenge through its unified domain approach. Compared to the second-best method, ConstStyle achieves improvements in mean average precision (\textit{mAP}) ranging from 0.1\% to 2.8\%.
%adopts a different approach by projecting unseen test instances onto known instances. 
% compared to the second-best method.
% This strategy leads to an improvement in mean average precision (mAP) ranging from 0.1\% to 2.8\% compared to the second-best method.

\begin{table}[t]
    \centering
    \resizebox{\linewidth}{!}{
        \begin{tabular}{l|cccc|cccc}
         \toprule
         \multicolumn{1}{c|}{\multirow{2}{*}{\textbf{Method}}} & \multicolumn{4}{c|}{\textbf{Market $\rightarrow$ Duke}} &  \multicolumn{4}{c}{\textbf{Duke $\rightarrow$ Market}} \\
         & mAP & R1 & R5 & R10 & mAP & R1 & R5 & R10 \\
         \midrule
         ERM & 13.8 & 27.1 & 40.5 & 46.9 & 21.9 & 45.9 & 65.6 & 72.7 \\
         MixStyle\cite{zhou2021domain} & 19.5 & 37.0 & 52.0 & 58.5 & 24.6 & 52.2 & 70.9 & 78.4 \\
         DSU\cite{li2022uncertainty} & 21.6 & 40.2 & 54.0 & 59.8 & \nduong{25.5} & \nduong{55.7} & \nduong{73.0} & \nduong{79.2} \\
         CSU\cite{zhang2024domain} & \nduong{24.1} & \nduong{44.2} & \nduong{59.2} & \nduong{65.1} & 24.9 & 55.0 & 72.2 & 78.2 \\
         \midrule
         \textbf{ConstStyle} & \nam{26.1} & \nam{44.7} & \nam{61.1} & \nam{67.6} & \nam{27.0} & \nam{56.5} & \nam{75.0} & \nam{81.1} \\
         \bottomrule
        \end{tabular}
    }
    \caption{Performance of the methods on Instance retrieval task. The best result is colored \nam{purple} and the second best result is colored \nduong{blue.}}
    \label{tab:instance_retrieval}
    \vspace{-10pt}
\end{table}

\section{Conclusion}

%\textcolor{red}{This part need to be rewrite and much more details and also provide an outlook for future work.}

This paper introduces ConstStyle, a novel approach for addressing the domain shift problem. The key concept of ConstStyle is to align data to a unified domain prior to both training and testing, enabling the capture of domain-invariant features and reducing discrepancies with unseen domains. This alignment approach mitigates domain shift effects and maintains performance stability, even with fewer training domains. ConstStyle consistently outperforms existing methods on style-shift datasets, achieving up to a 19.82\% accuracy improvement over the next best approach. Additionally, we provide a theoretical analysis on the performance bounds for both seen and unseen domains. Future work will explore addressing other types of domain shift, such as feature shift.
%However, it does not ensure robust performance across all domain shift types, such as feature shifts. A promising research direction is to integrate a feature-handling mechanism into ConstStyle to further enhance the performance on both style-shift and feature-shift datasets.
\section{Acknowledgement}
This research is funded by Hanoi University of Science and Technology (HUST) under grant number T2024-TD-002.

{
    \small
    \bibliographystyle{ieeenat_fullname}
    \bibliography{main}
}

\appendix
\clearpage
\setcounter{page}{1}
\maketitlesupplementary
\renewcommand{\thesection}{\Alph{section}}
\setcounter{section}{0}
\section{Details of the Training Process}
Details of unified domain determination algorithm, training and inference processes are presented in Algorithms \ref{algo:training}, \ref{algo:unified_domain} and \ref{algo:testing}, respectively.
\begin{algorithm*}[ht]
    \caption{ConstStyle Training Process}
    \label{algo:training}
    \textbf{Input:} Seen data $\mathcal{S} = \{(x, y)\}$, Model $\omega=\zeta(\theta_f(\theta_s(.))$, the update interval $\gamma$, the number of epochs $E$, the learning rate $\eta$, and the number of clusters $N'$\;
    \textbf{Output:} {Optimal model $\omega^*$, the final unified domain $\mathcal{N}^T$}\;
    \textbf{Algorithm:}\\
    \For{epoch $\leq$ $E$}
    {
        $\varepsilon \leftarrow \emptyset$\tcp*{Set of style features}
        \For{$x \in \mathcal{S}$}
        {
            \If{epoch $\leq \xi$}
            {
                $z_x = \theta_s(x)$\;
                $p(x) = \zeta(\theta_f(z_x))$\;
            }
            \Else
            {
                $z_x = \theta_s(x)$\;
                $\epsilon_s \sim \mathcal{N}^T$ \tcp*{sample style features}
                $\mu_s, \sigma_s=split(\epsilon_s)$\;
                $z_x^T = \sigma_s * \frac{z_x-\mu_x}{\sigma_x} + \mu_s$ \tcp*{project to the unified domain}
                $p(x) = \zeta(\theta_f(z_x^T))$\;
            }
            $l = \sum_{c\in \mathcal{C}} y_c.\log(p_c(x))$\;
                $\omega = \omega - \eta .\nabla_{\omega}l$\tcp*{Update model}
            \If{epoch \% $\gamma$ == 0}
            {   $\mu_{x_c} = \frac{1}{HW}\sum_{h=1}^H \sum_{w=1}^W z_{x_{c, h, w}},\sigma_{x_c} = \sqrt{\frac{1}{HW}\sum_{h=1}^H \sum_{w=1}^W (z_{x_{c, h,w}}-\mu_{x_c})^2}$\;
                $\epsilon_x =concat(\mu_x, \sigma_x)$ \tcp*{extract style features}
                $\varepsilon =\varepsilon \cup \epsilon_x$ \tcp*{store style features}
            }
        }
        \If{epoch \% $\gamma$ == 0}
        {
            % $\mathcal{M}=(\mathcal{C}_1, \mathcal{C}_2, ..., \mathcal{C}_{N'}) \leftarrow BayesGMM(\epsilon, N')$ \tcp*{clustering style of seen domains}
            % $\epsilon = \frac{1}{N'} \sum _{k=1}^{N'} \epsilon_{\mathcal{C}_k}$\;
            % $\Sigma = \frac{1}{N'} \sum _{k=1}^{N'} \Sigma_{\mathcal{C}_k}$\;
            $\mathcal{N}(\epsilon^T,\Sigma^T) = $Unified Domain Determination$(\varepsilon,N')$\; 
            $\mathcal{N}^T \leftarrow \mathcal{N}(\epsilon^T, \Sigma^T)$ \tcp*{get unified domain style}
        }
        $\omega^* = \omega$\;
    }
    \Return{$\omega^*, \mathcal{N}^T$}
\end{algorithm*}

\begin{algorithm}[t]
    \caption{Unified Domain Determination}
    \label{algo:unified_domain}
    \textbf{Input:} Set of all style features $\varepsilon=\{\epsilon_x|x\in S\}$, Number of clusters $N'$\;
    \textbf{Output:} Unified Domain Style $\mathcal{N}(\epsilon^T, \Sigma^T)$\;
    \textbf{Algorithm:}\\
    $\{\mathcal{C}_k \sim \mathcal{N}(\epsilon_{\mathcal{C}_k},\Sigma_{\mathcal{C}_k})|k=1..N'\} \leftarrow BayesGMM(\varepsilon, N')$ \
    $\epsilon^T = \frac{1}{N'} \sum _{k=1}^{N'} \epsilon_{\mathcal{C}_k}$\;
    $\Sigma^T = \frac{1}{N'} \sum _{k=1}^{N'} \Sigma_{\mathcal{C}_k}$\;
    $\mathcal{N}^T=\mathcal{N}(\epsilon^T, \Sigma^T)$\;
    \Return{$\mathcal{N}^T$}
\end{algorithm}

\section{Proofs}
% We begin with the Lemma 1.\\
% \textbf{Lemma 1.} With the target domain $\mathcal{T}$, the Seen risk of each source domains $\mathcal{S}_k \in \mathcal{S}$, denoted by $L^{{\mathcal{S}_k}^{T}}$ after converting to target domain $\mathcal{T} $ is bounded by the following inequation: 
% \begin{equation}
%     L^{\mathcal{S}_k^T} \leq L^{\mathcal{S}_k} + \beta \times( \mathcal{D}_\mu(\mathcal{T}, \mathcal{S}_k) +\mathcal{D}_\sigma(\mathcal{T},\mathcal{S}_k))
% \end{equation}

% where $\beta$ is the Lipschiz coefficient of loss function $l$ on all seen domains $\mathcal{S}$, $L^{\mathcal{S}_k}$ is the loss function of the model tested on seen domains by ERM.\\

\subsection{Proof of Lemma 1.}
\label{proof:lemma1}
Let us start with $L^{\mathcal{S}_k}$, we have:

\begin{align}
L^{\mathcal{S}_k} & =\frac{1}{|\mathcal{S}_k|}\sum_{(x, y) \in \mathcal{S}_k}[l(\omega(x), y)] \nonumber \\
& = \frac{1}{|\mathcal{S}_k|}\sum_{(x, y) \in \mathcal{S}_k}l(\zeta(\theta_f(z_x)), y) \nonumber \\
& = \frac{1}{|\mathcal{S}_k|}\sum_{(x, y) \in \mathcal{S}_k} l(\zeta(\theta_f(\sigma_x*\frac{z_x - \mu_x}{\sigma_x}+\mu_x)),y) \nonumber \\
& = \frac{1}{|\mathcal{S}_k|}\sum_{(x, y) \in \mathcal{S}_k}f(\mu_x, \sigma_x,\frac{z_x - \mu_x}{\sigma_x}, y).
\label{loss:l_S}
\end{align}

Similarly, we have:
\begin{align}
L^{\mathcal{S}^T_k} =& \frac{1}{|\mathcal{S}_k|}\sum_{(x, y) \in \mathcal{S}_k}[l(\omega^T(x), y)] \nonumber \\
=& \frac{1}{|\mathcal{S}_k|}\sum_{(x, y) \in \mathcal{S}_k}l(\zeta(\theta_f(z_x^T)), y) \nonumber \\
=& \frac{1}{|\mathcal{S}_k|}\sum_{(x, y) \in \mathcal{S}_k} l(\zeta(\theta_f(\sigma^T*\frac{z_x - \mu_x}{\sigma_x}+\mu^T)),y) \nonumber \\
=& \frac{1}{|\mathcal{S}_k|}\sum_{(x, y) \in \mathcal{S}_k}f(\mu^T, \sigma^T,\frac{z_x - \mu_x}{\sigma_x}, y).
\label{loss:l_S_delta}
\end{align}

By subtracting \ref{loss:l_S_delta} from \ref{loss:l_S}, we obtain: 
% \begin{equation}
%     L^{{\mathcal{S}_k}^{\Delta}} - L^{\mathcal{S}_k}= \frac{1}{|\mathcal{S}_k|}\sum_{(x,y) \in \mathcal{S}_k}(f(\mu^T, \sigma^T, z_{x_{norm}}, y) - f(\mu_{x},\sigma_{x}, z_{x_{norm}}, y)).
% \end{equation}
\begin{align}
    L^{\mathcal{S}^T_k} - L^{\mathcal{S}_k} &= 
    \frac{1}{|\mathcal{S}_k|} \sum_{(x,y) \in \mathcal{S}_k} \Big( f(\mu^T, \sigma^T, \frac{z_x - \mu_x}{\sigma_x}, y) \notag \\
    &\quad - f(\mu_{x}, \sigma_{x}, \frac{z_x - \mu_x}{\sigma_x}, y) \Big).
\end{align}

Using the Taylor approximation for a function with two variables, we derive:
% \begin{equation}
% \begin{split}
%     f(\mu^T, \sigma^T, z_{x_{norm}}, y) - f(\mu_{x},\sigma_{x}, z_{x_{norm}}, y) & 
%     \approx (\mu^T - \mu_{x})*\nabla_{\mu_{x}} f + (\sigma^T - \sigma_{x})*\nabla_{\sigma_{x}} f  
%     % \\
%     % & = (\mu^T - \mu_{x})*\nabla_{\mu_{x}} f + (\sigma^T - \sigma_{x})*\nabla_{\sigma_{x}} f
% \end{split}
% \end{equation}

\begin{multline}
    f(\mu^T, \sigma^T, \frac{z_x - \mu_x}{\sigma_x}, y) - f(\mu_{x}, \sigma_{x}, \frac{z_x - \mu_x}{\sigma_x}, y) \\
    \approx (\mu^T - \mu_{x}) \cdot \nabla_{\mu_{x}} f 
    \quad + (\sigma^T - \sigma_{x}) \cdot \nabla_{\sigma_{x}} f.
\end{multline}

Let $\mathcal{D}_\mu(\mathcal{T}, \mathcal{S}_k)$ denote the distance between means of the unified instance style $\mathcal{T}$ and seen instance style $\mathcal{S}_k$, while $\mathcal{D}_\sigma(\mathcal{T}, \mathcal{S}_k)$ represents the distance between standard deviations. Let $||v||$ denote the L2-norm of vector $v$. Assume $f$ is a $\beta$-Lipschitz function, we can suppose $\sup_{x \in \mathcal{S}_k}||\nabla_{\mu_{x}}f|| =\beta_\mu$, $\sup_{x \in \mathcal{S}_k}||\nabla_{\sigma_{x}}f|| = \beta_\sigma$, we have:

\begin{align*}
    \begin{split}
        L^{\mathcal{S}_k^T} - L^{\mathcal{S}_k} \leq & 
        \frac{1}{|\mathcal{S}_k|}\sum _{(x, y) \in \mathcal{S}_k} (\beta_\mu*||\mu^T - \mu_x|| + \\
        & \beta_\sigma*||\sigma^T - \sigma_x||)
    \end{split}
    \\
    % \begin{split}
    %     = & \frac{1}{|\mathcal{S}_k|}\sum_{(x, y) \in \mathcal{S}_k} \beta_\mu*||\mu^T - \mu_x|| + \\
    %     & \beta_\sigma||\sigma^T - \sigma_x||
    % \end{split}
    % \\
    \begin{split}
        = & \beta_\mu *\frac{1}{|\mathcal{S}_k|}\sum_{(x, y) \in \mathcal{S}_k} ||\mu^T - \mu_x|| + \\
        & \beta_\sigma *\frac{1}{|\mathcal{S}_k|}\sum_{(x, y) \in \mathcal{S}_k} ||\sigma^T - \sigma_x||
    \end{split}
    \\
    \begin{split}
        = & \beta_\mu * \mathcal{D}_\mu(\mathcal{T}, \mathcal{S}_k) + \beta_\sigma*\mathcal{D}_\sigma(\mathcal{T},\mathcal{S}_k)
    \end{split}
\end{align*}

Let $\beta = \max(\beta_\mu, \beta_\sigma)$, then:
\begin{align}
    \begin{split}
        L^{\mathcal{S}_k^T} - L^{\mathcal{S}_k} \leq \beta \times (\mathcal{D}_\mu(\mathcal{T}, \mathcal{S}_k) +\mathcal{D}_\sigma(\mathcal{T},\mathcal{S}_k))
    \end{split}
\end{align}

\subsection{Proof of Theorem 1}
\label{proof:theorem1}
According to \ref{proof:lemma1}, for the seen domains $\{\mathcal{S}_k\}_{k=1}^{N}$, the total empirical loss across $N$ seen domains is bounded as follows:

\begin{multline}
    \sum_{k=1}^N L^{{\mathcal{S}_k}^T} \leq \sum_{k=1}^N L^{\mathcal{S}_k} + \beta*\sum_{k=1}^N(\mathcal{D}_\mu(\mathcal{T},\mathcal{S}_k) + \mathcal{D}_\sigma(\mathcal{T},\mathcal{S}_k))
\end{multline}

It can be observed that the upper bound of this loss depends on the total distance from the unified domain to $N$ seen domains $\mathcal{S}_k$. 
Therefore, to minimize the loss over the seen domains, we aim to reduce the distance between the unified domain $\mathcal{T}$ and $N$ seen domains $\mathcal{S}_k$. 
Consequently, the unified domain style $\mathcal{N}^T = (\mu^T,\Sigma^T)$ is the barycenter of $N$ seen domain styles.

\subsection{Proof of Theorem 2}
\label{proof:theorem2}
The loss function of the model trained on seen domains, obtained by ConstStyle, and test on unseen domain is given by:

\begin{align}
L^{\mathcal{U}^T} & =\frac{1}{|\mathcal{U}|}\sum_{(u, y) \in \mathcal{U}}[l(\omega^T(u), y)] \nonumber \\
& = \frac{1}{|\mathcal{C}|}\sum_{c \in \mathcal{C}} \frac{1}{|\mathcal{U}_c|} \sum_{u \in \mathcal{U}_c}[l(\zeta(\theta_f(z_u^T))), y_c] \nonumber \\
& = \frac{1}{|\mathcal{C}|}\sum_{c \in \mathcal{C}}\frac{1}{|\mathcal{U}_c|}\sum_{u \in \mathcal{U}_c}[l(\zeta(\theta_f(\sigma_u^T*\frac{z_u - \mu_u}{\sigma_u}+\mu_u^T),y_c] \nonumber \\
& = \frac{1}{|\mathcal{C}|}\sum_{c \in \mathcal{C}}\frac{1}{|\mathcal{U}_c|}\sum_{u \in \mathcal{U}_c}[f(\mu_u^T,\sigma_u^T,\frac{z_u - \mu_u}{\sigma_u},y_c)]
\label{loss:l_u}
\end{align}

Similarly, We have:

\begin{align}
        L^{\mathcal{S}^T} & =\frac{1}{|\mathcal{S}|}\sum_{(x, y) \in \mathcal{S}}[l(\omega^T(x), y)] \nonumber \\
        & = \frac{1}{|\mathcal{C}|}\sum_{c \in \mathcal{C}}\frac{1}{|\mathcal{S}_c|}\sum_{x \in S_c}[l(\zeta(\theta_f(z^T_x))),y_c] \nonumber \\
        & = \frac{1}{|\mathcal{C}|}\sum_{c \in \mathcal{C}}\frac{1}{|\mathcal{S}_c|}\sum_{x \in \mathcal{S}_c}[l(\zeta(\theta_f(\sigma^T*\frac{z_x - \mu_x}{\sigma_x}+\mu^T),y_c] \nonumber \\
        & = \frac{1}{|\mathcal{C}|}\sum_{c \in \mathcal{C}}\frac{1}{|\mathcal{S}_c|}\sum_{x \in \mathcal{S}_c}[f(\mu^T,\sigma^T,\frac{z_x - \mu_x}{\sigma_x},y_c)].
        \label{loss:l_u_delta}
\end{align}

% \vspace{10pt}
Assume that the cardinality of seen domain $S$ and unseen domain $U$ are the same for all classes, i.e, $|\mathcal{S}_c|=|\mathcal{U}_c|=d=\frac{|\mathcal{U}|}{|\mathcal{C}|}=\frac{|\mathcal{S}|}{|\mathcal{C}|},\forall c \in \mathcal{C}$ . 
From Equations \ref{loss:l_u} and \ref{loss:l_u_delta}, we have: 

\begin{multline}
    L^{\mathcal{U}^T} - L^{\mathcal{S}^T} = \frac{1}{|\mathcal{C}|}\sum_{c\in \mathcal{C}}\frac{1}{d}\sum_{u \in U_c, x \in S_c} [f(\mu_u^T,\sigma_u^T,\frac{z_u - \mu_u}{\sigma_u},y_c)\\
    -f(\mu^T, \sigma^T, \frac{z_x - \mu_x}{\sigma_x},y_c)]
\end{multline}
By applying the Taylor approximation for three variables, we obtain:

\begin{algorithm}[t]
    \caption{ConstStyle Inference Process}
    \label{algo:testing}
    \textbf{Input:} {Unseen data $\mathcal{U}= \{u| u\sim\mathcal{U}\}$, Optimal model $\omega^*$, Unified domain $\mathcal{N}^T$}\;
    \textbf{Output:} Prediction set $L_\mathcal{U}$\;
    \textbf{Algorithm:}\\
    $L_\mathcal{U}=\emptyset$\;
    \For{$u \in \mathcal{U}$}
    {
        $z_u = \theta_s(u)$\;
        $\mu^T, \sigma^T=split(\epsilon^T)$\; 
        $ z_u^T = (\alpha.\sigma_u + (1 - \alpha).\sigma^T) . \frac{z_u-\mu_u}{\sigma_u} + (\alpha.\mu_u + (1 - \alpha).\mu^T)$\;
        $p(u) = \zeta(\theta_f(z_u^T))$\;
        $y_u = \arg \max(softmax(p(u))$\;
        $L_{\mathcal{U}} = L_{\mathcal{U}} \cup y_u$\;
    }
    \Return{$L_{\mathcal{U}}$}
\end{algorithm}

\begin{align}
&f(\mu_u^T, \sigma_u^T, \frac{z_u - \mu_u}{\sigma_u}, y_c) - f(\mu^T, \sigma^T, \frac{z_x - \mu_x}{\sigma_x}, y_c) \nonumber \\
&\approx (\mu_u^T - \mu^T)\nabla_{\mu^T} f + (\sigma_u^T - \sigma^T)\nabla_{\sigma^T}f \nonumber \\
&\quad + (\frac{z_u - \mu_u}{\sigma_u}-\frac{z_x - \mu_x}{\sigma_x})\nabla_{\frac{z_x - \mu_x}{\sigma_x}}f \nonumber \\
&= (\alpha * \mu_u + (1 - \alpha) * \mu^T - \mu^T)\nabla_{\mu^{T}} f \nonumber \\
&\quad + (\alpha * \sigma_u + (1 - \alpha) * \sigma^T - \sigma^T)\nabla_{\sigma^T} f \nonumber \\
&\quad + (\frac{z_u - \mu_u}{\sigma_u}-\frac{z_x - \mu_x}{\sigma_x})\nabla_{\frac{z_x - \mu_x}{\sigma}}f \nonumber \\
&= \alpha * (\mu_u - \mu^T)*\nabla_{\mu^T} f + \alpha * (\sigma_u - \sigma^T)*\nabla_{\sigma^T} f \nonumber \\
&\quad + (\frac{z_u - \mu_u}{\sigma_u}-\frac{z_x - \mu_x}{\sigma})\nabla_{\frac{z_x - \mu_x}{\sigma}}f
\label{long_equation}
\end{align}

Denote $||v||$ as the L2-norm of tensor $v$. Suppose that $\sup_{x \in \mathcal{S}} (||\nabla_{\mu^T}f||, ||\nabla_{\sigma^T}f||) = \beta$ and $\sup_{x \in S}\nabla_{\frac{z_x - \mu_x}{\sigma}}f=\xi$, then we have:

\begin{align}
    \label{loss:notdone}
    \nonumber
    & L^{\mathcal{U}^T} - L^{\mathcal{S}^T} \\ 
    \nonumber
    &\leq  
    \frac{1}{|\mathcal{C}|}\sum_{c \in \mathcal{C}}\frac{1}{d}\sum _{u \in \mathcal{U}_c, x \in \mathcal{S}_c} (\alpha \times (\beta \times||\mu^T - \mu_u|| \\
    \nonumber
        & \quad + \beta \times||\sigma^T - \sigma_u||)+\xi \times ||\frac{z_u - \mu_u}{\sigma_u}-\frac{z_x - \mu_x}{\sigma}||)
    \\
    \nonumber
    & \leq \alpha \times \beta \times\frac{1}{|\mathcal{U}|}\sum_{u \in \mathcal{U}} ( ||\mu^T - \mu_u||+||\sigma^T - \sigma_u||) \\
    & \quad + \xi \times \frac{1}{|\mathcal{U}|}\sum_{u \in \mathcal{U}, x \in \mathcal{S}}||\frac{z_u - \mu_u}{\sigma_u}-\frac{z_x - \mu_x}{\sigma}||.\\
    \nonumber
\end{align}

Observed that, $\frac{z_u - \mu_u}{\sigma_u},\frac{z_x - \mu_x}{\sigma}\sim \mathcal{N}(0,I)$, where $I$ is the identity matrix size $C \times H \times W$, where $C, H, W$ are the channel, height, and width dimensions of $z_x$. When the cardinality of seen domains $\mathcal{S}$, unseen domain $\mathcal{U}$ is sufficiently large, we can approximate:
\begin{equation}
    \frac{1}{|\mathcal{U}|}\sum_{u \in \mathcal{U}, x \in \mathcal{S}}||\frac{z_u - \mu_u}{\sigma_u}-\frac{z_x - \mu_x}{\sigma}|| = \mathbb{E}[||U-X||],
\end{equation}

where $U$ and $X$ are two random multivariate variables over $\mathbb{R}^{C\times H\times W}$ drawn from standard Gaussian distribution, $U, X \sim \mathcal{N}(0, I)$. 
We have:
\begin{align*}
    &\quad \quad0 \leq \mathbb{V}[||U - X||] = \mathbb{E}[||U-X||^2] - (\mathbb{E}[U - X])^2 \\
    & \rightarrow \mathbb{E}[U-X] \leq \sqrt{\mathbb{E}[||U-X||^2]} = \sqrt{Tr(2I)} \\
    & \rightarrow \frac{1}{|\mathcal{U}|}\sum_{u \in \mathcal{U}, x \in \mathcal{S}}||\frac{z_u - \mu_u}{\sigma_u}-\frac{z_x - \mu_x}{\sigma}|| \leq \sqrt{Tr(2I)}
\end{align*}

Let $\mathcal{D}_\mu(\mathcal{T}, \mathcal{U})$ and $\mathcal{D}_\sigma(\mathcal{T}, \mathcal{U})$ be the distance between mean and standard deviation of unified domain $\mathcal{T}$ and unseen domain $\mathcal{U}$, respectively. From Equation ~\eqref{loss:notdone}, we obtain:
\begin{align*}
    L^{\mathcal{U}^T}-L^{\mathcal{S}^T}& \leq \alpha \times \beta \times (\mathcal{D}_\mu(\mathcal{U}, \mathcal{T})+\mathcal{D}_\sigma(\mathcal{\mathcal{U}, \mathcal{T}})) \\
    & \quad +\xi \times \sqrt{2.Tr(I)}
\end{align*}

\begin{table*}[t]
    \centering
    \resizebox{1.0\linewidth}{!}{
        \setlength{\tabcolsep}{7pt}
        \begin{tabular}{l|c|cccccccccc|r}
         \toprule
         Method & Venue & M,MM & M,S & M,SY & M,U & MM,S & MM,SY & MM,U & S,SY & S,U & SY,U & Avg \\
         \midrule
         ERM & - & \nam{80.22} & 82.77 & 92.34 & 97.46 & 77.60 & 74.83 & 71.67 & 52.49 & 77.70 & 87.86 & 79.49 \\
         Crossgrad & ICLR 2018 & 79.24 & 82.95 & 92.01 & \nduong{97.68} & 76.64 & 75.01 & 73.00 & 50.77 & 78.77 & 84.8 & 77.02 \\ 
         Mixup & ICLR 2018 & 75.92 & \nduong{84.88} & 90.81 & 96.75 & 75.87 & 70.71 & 67.49 & 44.03 & \nduong{80.51} & 82.58 & 76.95 \\ 
         Cutmix & ICCV 2019 & 74.86 & \nam{85.16} & 91.61 & 97.02 & 77.78 & 70.04 & 68.87 & 45.51 & \nam{80.75} & 85.59 & 77.71 \\
         EFDMix & CVPR 2022 & 76.29 & 82.87 & 92.53 & 97.52 & 77.65 & \nduong{76.14} & \nduong{73.33} & 52.34 & 78.57 & 85.87 & 78.88 \\ 
         RIDG & ICCV 2023 & 79.75 & 84.48 & 91.97 & 97.23 & 77.8 & 73.77 & 71.05 & 50.73	& 79.74	& 86.33	& 79.28 \\
         MixStyle & ICLR 2021 & 77.96 & 72.69 & 83.37 & 86.82 & 75.09 & 62.18 & 68.15 & 41.53 & 58.5 & 71.88 & 69.81 \\
         DSU & ICLR 2022 & 78.77	& 83.83	& 92.1 & \nam{97.81} & \nduong{78.53} & 74.78 & 71.89 & 53.66 & 78.14 & 87.62 & 79.71 \\
         CSU & WACV 2024 & 78.64 & 84.29 & \nduong{92.72} & 97.39 & 77.27 & 75.61 & 72.67 & \nduong{57.28} & 78.56 & \nduong{88.08} & \nduong{80.25}\\ 
         ConstStyle & Ours & \nam{80.22} & 84.69 & \nam{92.92} & 97.33 & \nam{78.73} & \nam{76.27} & \nam{74.19} & \nam{57.58} & 80.29 & \nam{88.24} & \nam{81.04} \\
         \midrule
         Method & Venue & M,MM,S & M,MM,SY & M,MM,U & M,S,SY & M,S,U & M,SY,U & MM,S,SY & MM,S,U & MM,SY,U & S,SY,U & Avg \\
         \midrule
         ERM & - & 80.12 & \nduong{76.39} & \nam{71.41} & 61.89 & 83.47 & 91.07 & 44.49 & 77.63 & \nduong{76.87} & 48.47 & 71.18\\
         Crossgrad & ICLR 2018 & 79.59 & 76.32 & \nduong{71.31} & 60.37 & 83.21 & 91.47 & 36.57 & 77.71 & 74.26 & 46.55 & 70.34 \\ 
         Mixup & ICLR 2018 & 78.35 & 74.19 & 69.51 & 57.22 & \nam{85.78} & 91.16 & 34.45 & 77.11 & 71.34 & 41.29 & 68.04 \\ 
         Cutmix & ICCV 2019 & 79.82 & 73.12 & 68.92 & 58.28 & \nduong{85.64} & 91.32 & 32.52 & \nduong{78.3} & 72.57 & 39.92 & 68.04 \\
         EFDMix & CVPR 2022 & 80.38 & 76.04 & 70.13 & 63.48 & 83.62 & \nduong{91.96} & 43.46 & 77.61 & 73.94 & \nduong{50.43} & 71.10 \\ 
         RIDG & ICCV 2023 & 80.51 & 74.71 & 70.45 & 61.76 & 84.78 & 91.41 & 35.02 & 78.28 & 75.74 & 45.53 & 69.81 \\
         MixStyle & ICLR 2021 & 78.91 & 74.97 & 61.48 & 57.95 & 71.44 & 81.43 & 42.92 & 71.44 & 62.3 & 40.99 & 64.38 \\
         DSU & ICLR 2022 & \nam{80.71} & 76.25 & 70.54 & 62.35 & 83.25 & 91.47 & 42.87 & 77.84 & 76.29 & 48.31 & 70.98 \\
         CSU & WACV 2024 & \nduong{80.63} & 76.26 & 69.50 & \nam{64.68} & 85.09 & 91.53 & \nduong{47.31} & 77.64 & 75.61 & 52.73 & \nduong{72.09} \\ 
         ConstStyle & Ours & 80.32 & \nam{77.93} & 70.89 & \nam{64.68} & 84.88 & \nam{92.10} & \nam{48.88} & \nam{79.08} & \nam{77.27} & \nam{53.55} & \nam{72.95} \\
         \bottomrule
        \end{tabular}
    }
    \caption{Multiple unseen domain generalization (2 and 3 unseen domains) on Digits5 dataset. Abbrevations: (M: MNIST, MM: MNISTM, S: SVHN, SY: SYN, U: USPS). The best result is colored \nam{purple} and the second best result is colored \nduong{blue}.}
    \label{tab:digits2}
\end{table*}

\section{Experiment Setup}
\label{apdx:setup}
\textbf{Image classification}: We train a ResNet18 pretrained on ImageNet for 200 epochs with learning rate of 0.001. Batch size is set to 32 for PACS dataset with 3 integrated ConstStyle layers, and 128 with 1 ConstStyle layer for Digit5 dataset. \\
\textbf{Image Corruption}: We use WideResNet with a single ConstStyle layer as a backbone, training for 200 epochs with a learning rate of 0.05 and batch size of 512. \\
\textbf{Instance Retrieval}: We train a model with ResNet50 pretrained on ImageNet as the backbone for 80 epochs with a learning rate of 0.0035. We integrate 3 ConstStyle layers into the model.\\
Across all experiment scenarios, the number of clusters is fixed to 4. All methods are optimized using SGD optimizer. Optimal hyperparameters are selected based on the performance on the validation dataset.

\begin{figure}[t]
    \centering
    \includegraphics[width=1.0\linewidth]{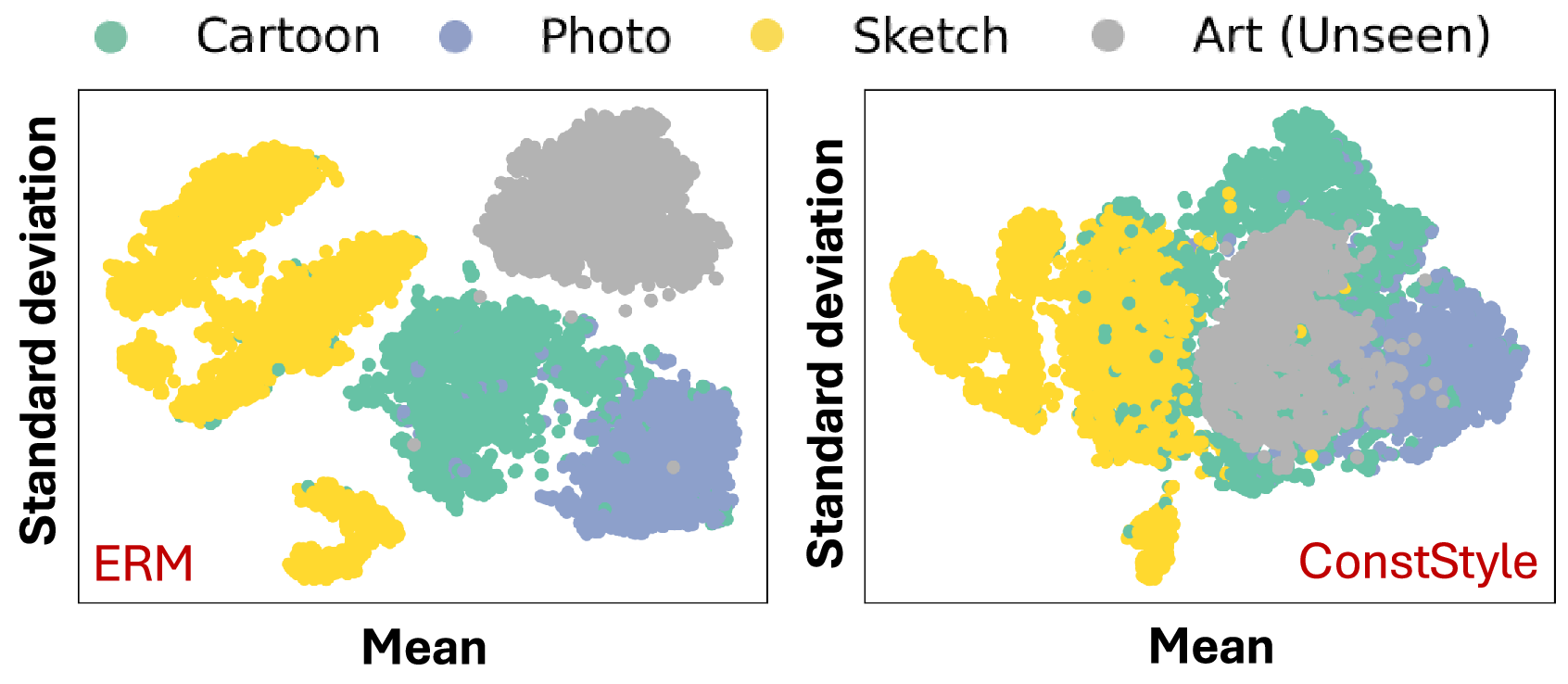}
    \captionof{figure}{Style statistics of ERM and ConstStyle.}
    \label{fig:features}
\end{figure}

\begin{table}[t]
    \centering
    \resizebox{0.8\linewidth}{!}{%
        \begin{tabular}{l|cc}
            \toprule
            \multicolumn{1}{c|}{\multirow{2}{*}{\textbf{Method}}} & \multicolumn{2}{c}{\textbf{Dataset}} \\ 
            \cmidrule{2-3}
            & PACS & Digit5 \\ 
            \midrule
            ConstStyle w/ Pretrained features & 86.31 & 76.61 \\
            ConstStyle w/ Domain label & 86.73 & 86.37 \\
            ConstStyle & \nam{86.77} & \nam{86.88} \\
            \bottomrule
        \end{tabular}
    }
    \caption{Different variants of ConstStyle.}    
    \label{tab:conststyle_results}
\end{table}

\section{Additional Results}

\subsection{Multiple Unseen Domains on Digit5 dataset}
\label{x:digit5_multi_unseen}
We perform additional experiment with multiple unseen domains on the Digit5 dataset.
The results are shown in Table \ref{tab:digits2}. It can be observed that ConstStyle achieves the best performance in most of the scenarios, and obtains the highest average accuracy.

\section{Ablation Studies}
In this section, we conduct a more in-depth analysis concerning the impacts hyperparameters in ConstStyle's, which is the number of clusters used during the unified domain determination phase, we additionally perform experiments to explore the influence of training batch size and impact of $\alpha$ in the inference process.

\begin{table}[t]
    \centering
    \resizebox{0.8\linewidth}{!}{%
        \begin{tabular}{l|ccccc}
             \toprule
             \# of clusters & 1 & 2 & 3 & 4 & 5 \\
             \midrule
             PACS & 86.07 & 86.51 & 86.61 & \nam{86.77} & 86.61 \\
             Digit5 & 85,80 & 85,60 & 85,54 & \nam{86.88} & 85.93 \\
             \bottomrule
        \end{tabular}
    }
    \caption{Impacts of the number of clusters.}
    \label{fig:abs1}
\end{table}

\begin{table}[t]
    \centering
    \resizebox{0.8\linewidth}{!}{
        \begin{tabular}{l|cccccc}
         \toprule
         Batchsize & 8 & 16 & 32 & 64 & 128 & 256 \\
         \midrule
         Accuracy & 85.43 & 86.22 & \nam{86.77} & 86.33 & 85.91 & 85.10 \\
         \bottomrule
        \end{tabular}
    }
    \caption{Impacts of the batch size on accuracy (PACS dataset).}
    \label{tab:ab2}
\end{table}

\subsection{In-depth analysis of ConstStyle}
We first conduct additional experiments to further analyze the behaviors of ConstStyle. Figure \ref{fig:features} illustrates the style statistics for both seen and unseen domains, demonstrating that ConstStyle effectively aligns training and test samples within a unified domain, thereby enhancing performance under distribution shift.
Additionally, we evaluate ConstStyle with two alternative approaches: \textbf{1. Clustering using domain label} and \textbf{2. Utilizing pretrained style statistics} with results shown in Table \ref{tab:conststyle_results}. We can observe that while domain labels can produce good performance, they are not always optimal, as some samples have style statistics belonging to other domains; thus, clustering using GMM can form appropriate domain clusters, yielding better performance.
Furthermore, using pretrained features for clustering can achieve comparable results if style features are previously learned by the pretrained model, as shown in the PACS dataset in Table \ref{tab:conststyle_results}. However, if the pretrained model has not learned style features, relying on them can significantly degrade ConstStyle's accuracy, as observed in the Digit5 dataset.

\begin{table}[t]
    \centering
    \resizebox{1.0\linewidth}{!}{
    \begin{tabular}{l|ccccccccccc}
         \toprule
         $\alpha$ & 0 & 0.1 & 0.2 & 0.3 & 0.4 & 0.5 & 0.6 & 0.7 & 0.8 & 0.9 & 1.0 \\
         \midrule
         PACS & 86.00 & 85.08 & 85.76 & 86.34 & 86.33 & 86.34 & \nam{86.77} & 86.46 & 86.22 & 86.03 & 85.86 \\
         Digit5 & 85.96 & 85.62 & 85.77 & 85.91 & 85.95 & \nam{86.88} & 86.01 & 85.99 & 85.98 & 86.01 & 85.96 \\
         \bottomrule
    \end{tabular}
    }
    \caption{Impact of $\alpha$ to the model performance on different datasets.}
    \label{tab:ab3}
\end{table}

\subsection{Impacts of the Number of Clusters }
We first investigate the impact of the number of clusters during the clustering phase, ranging from one to five. Figure \ref{fig:abs1} demonstrates that ConstStyle performs consistently across domains, regardless of the number of clusters. 
This consistency demonstrates ConstStyle's robustness, as the major goal is to construct a single domain by averaging all of the clusters in the visible domains. 

% \subsection{Impacts of the Number of Style Feature Extractor's Layers} 
% We next study how the number of the style feature extractor's layers affects ConstStyle's performance. Specifically, we vary the number of this parameter from 1 to 5, integrating them into each block of the ResNet18 model. 
% The results presented in Figure \ref{fig:abs2} reveal that integrating ConstStyle layers into all blocks responsible for extracting style attributes (blocks 1 to 4) enhances performance. 
% However, integrating a layer into block 5, which primarily extracts semantic features, introduces confusion about the actual class of the data, leading to a significant performance drop.

\subsection{Impacts of the Batch Size}
In this section, we investigate the impacts of the batch size on ConstStyle's performance. Experiments are conducted with batch size ranging from 8 to 256, and the results are presented in Table \ref{tab:ab2}. 
The results suggest that using either very small or very large batch sizes can lead to suboptimal performance, as too few or too many style modifications may disrupt learning stability. 
The optimal strategy is to use a moderate batch size (about $32$), ensuring balanced and steady learning for the model.

\subsection{Impacts of Partial Projection}
\label{appendix:alpha}
We study the impacts of $\alpha$ on the performance of the proposed method by varying this parameter from $0$ to $1$, with the results presented in Table \ref{tab:ab3}.
It is evident that the impact of $\alpha$ varies significantly across different values, highlighting its important role in achieving optimal performance. 
When an appropriate value of $\alpha$ is selected, overall performance can improve by up to $0.56\%$ for PACS dataset and up to $0.87\%$ for Digit5 dataset, compared to when no $\alpha$ value is used. This results also highlights the effects of our proposed partial style alignment algorithm (Section 3.5). 

\begin{table}[t]
    \centering
    \resizebox{1.0\linewidth}{!}{
        \begin{tabular}{c|llll}
            \toprule
            Data size & 32892 & 65787 & 98680 & 131575 \\
            \midrule
            Average training time per epoch (s) & 249.2 & 568.4 & 857.3 & 1076.4 \\
            \bottomrule
        \end{tabular}
    }
    \caption{Scalability of ConstStyle with different number of training data size.}
    \label{tab:scalability}
\end{table}

\subsection{Scalability against larger datasets}
ConstStyle has three components: style statistics distribution estimation, unified style determination, and style aligment. The computational complexity of all three components scales linearly with the training data size. 
As a result, Constlyle is inherently scalable to large datasets. This scalability is also empirically demonstrated in Table \ref{tab:scalability}, which reports the average training time per epoch when varying the training data size.

\end{document}